\documentclass{article}

     \usepackage[preprint]{neurips_2019}

\usepackage[utf8]{inputenc} %
\usepackage[T1]{fontenc}    %
\usepackage{hyperref}       %
\usepackage{url}            %
\usepackage{booktabs}       %
\usepackage{amsfonts}       %
\usepackage{nicefrac}       %
\usepackage{microtype}      %
\usepackage[title]{appendix}

\usepackage{graphicx}
\usepackage{subcaption}
\usepackage{natbib}
\usepackage{placeins}
\usepackage[normalem]{ulem} %
\usepackage{xspace}         %
\usepackage{amsmath}
\usepackage[colorinlistoftodos,prependcaption,textsize=small,linecolor=red,backgroundcolor=red!25,bordercolor=red]{todonotes}  %
\usepackage[export]{adjustbox}  %
\usepackage{verbatim}  %
\usepackage{wrapfig}  %
\usepackage{siunitx}  %
\usepackage{float}  %

\newcommand{\oldmodelname}{SPIRAL\xspace}
\newcommand{\newmodelname}{SPIRAL++\xspace}
\newcommand{\generativeagentname}{generative agent\xspace}
\newcommand{\generativeagentsname}{generative agents\xspace}
\newcommand{\xx}{{\mathbf x}}

\newcommand{\xxsub}[1]{\mathbf{x}_{\mbox{\scriptsize #1}}}
\newcommand{\mm}{{\mathbf m}}

\newcommand*{\eg}{{\em e.g.}\@\xspace}

\makeatletter
\newcommand*{\etc}{%
    \@ifnextchar{.}%
        {etc}%
        {etc.\@\xspace}%
}
\makeatother

\newcommand{\inlinesubsection}[1]{%
  \par
  \pagebreak[2]%
  \refstepcounter{subsection}%
    \everypar={%
      {\setbox0=\lastbox}%
      \addcontentsline{toc}{subsection}{%
        \thesubsection. \hspace*{3pt}#1}%
      \textbf{\thesubsection\space\space{#1}. }%
      \everypar={}%
    }%
  \ignorespaces
}

\title{Unsupervised Doodling and Painting \\ with Improved SPIRAL}

\author{%
John F. J. Mellor \And Eunbyung Park \And Yaroslav Ganin \And Igor Babuschkin \And Tejas Kulkarni \And Dan Rosenbaum \And Andy Ballard \And Theophane Weber \And Oriol Vinyals \And S. M. Ali Eslami\\
}

\begin{document}

\maketitle
 
\begin{abstract}
We investigate using reinforcement learning agents as generative models of images \citep{ganin2018spiral}. A \textit{\generativeagentname} controls a simulated painting environment, and is trained with rewards provided by a discriminator network simultaneously trained to assess the realism of the agent's samples, either unconditional or reconstructions. Compared to prior work, we make a number of improvements to the architectures of the agents and discriminators that lead to intriguing and at times surprising results. We find that when sufficiently constrained, \generativeagentsname can learn to produce images with a degree of visual abstraction, despite having only ever seen real photographs (no human brush strokes). And given enough time with the painting environment, they can produce images with considerable realism. These results show that, under the right circumstances, some aspects of human drawing can emerge from simulated embodiment, without the need for external supervision, imitation or social cues. Finally, we note the framework's potential for use in creative applications.
\end{abstract}

\begin{figure}[h]
\centering
    \includegraphics[width=\textwidth]{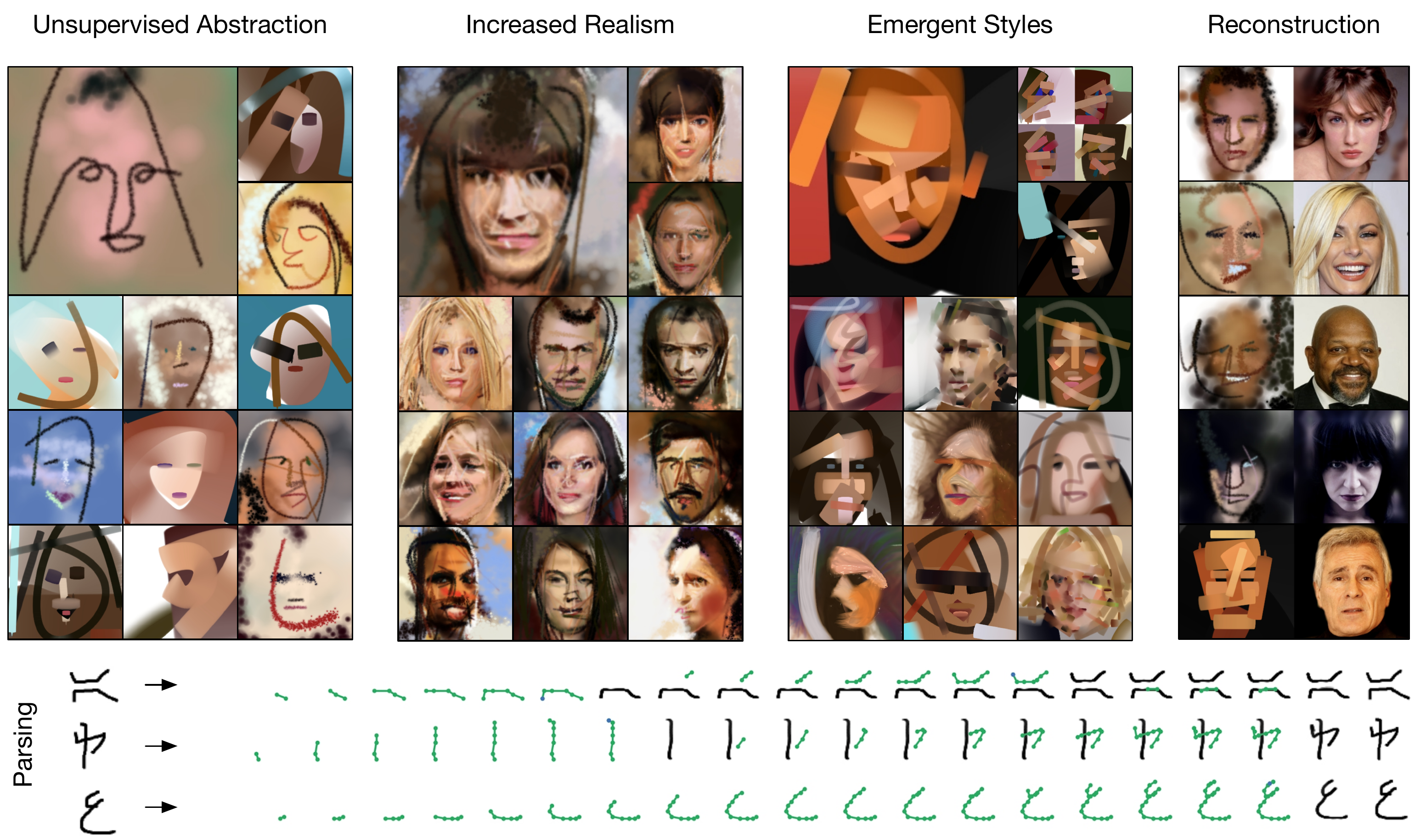}
\caption{Despite not seeing examples of human drawings, generative agents can produce images with a degree of visual abstraction, in a diverse array of styles, and they can scale to approach realistic results. Agents can also learn to reconstruct, for instance parsing Omniglot characters into strokes.}
\label{fig:hero}
\end{figure}

\section{Introduction}

Through a human's eyes, the world is much more than just the patterns of photons reflected in our retinas. We have the ability to form high-level interpretations of objects and scenes, and we use these to reason, to build memories, to communicate, to plan and to take actions.

An important open problem in artificial intelligence is how these interpretations are produced in the absence of labelled datasets to train from. The idea of generative modelling has offered a possible solution, whereby high-level representations are obtained by first training models to generate outputs that resemble real images \citep{kingma2013auto, gregor2015draw, eslami2018neural}. The underlying hypothesis is that if a concise representation is sufficient for the model to reproduce an image, then that representation has, in some form, captured the essence of the contents of that image.

Modern implementations of the generative modeling approach use large, flexible, mostly unstructured neural networks to model the generative process, with state-of-the-art models now achieving photo-realistic results \citep{karras2018style, brock2018biggan} even at high resolutions. Whilst immensely impressive, such results raise two intriguing questions: 1. To what extent is photo-realism necessary for learning of high-level descriptions of images? 2. Is it ever beneficial to incorporate grounded \textit{knowledge} and \textit{structure} in the generative model?

Humans have been representing and reconstructing their visual sensations, using physical tools to draw and sculpt depictions of those sensations, for at least 60,000 years \citep{hoffmann2018u}, well before the development of symbolic writing systems. And it has been hypothesized that abstraction away from raw sensations and use of physical tools are essential components of certain human cognitive abilities \citep{clark1998extended, lake2015human, lake2017building}. Within the field of artificial intelligence itself, human figurative drawings have long provided inspiration for the study of meaningful representation of object concepts \citep{minsky1972artificial}.

We would like to be able to create generative models that similarly use physical grounding. In this work, we equip artificial agents with the same tools that we use to create reconstructions of images (namely digital brushes, pens and spray cans). We train these agents with reinforcement learning to interact with digital painting environments \citep{renold2004mypaint,li2017fluid}.
The agents act by placing strokes on a simulated canvas and changing the brush size, pressure and colour as they do so. Building on the work by \cite{ganin2018spiral}, we consider a setting where the agents' rewards are specified by jointly-trained adversarial discriminator networks. An important aspect of this approach is that the structure is not in the agent itself, but mostly in the environment that it interacts with. 

In summary, we make the following contributions: \textbf{(a) Quality:} We scale up and tune the work of \cite{ganin2018spiral}. We use several simple but general tricks that improve the performance of the framework and show that these changes positively impact the fidelity and realism of generated images. \textbf{(b) Abstraction:} We demonstrate that \generativeagentsname can learn to draw recognizable objects with a small number of strokes of the simulated brush and without supervision. These drawings appear to capture the essential elements of objects at a surprisingly high level: for instance generating faces by drawing two eyes, a nose and a mouth each with a single stroke. We also show how the discriminators learn distance metrics that can prioritize semantic likeness over pixel similarity. Our results in this work are primarily qualitative, in part due to the subjective nature of the phenomena being studied, however we provide systematic ablations of the results in the supplementary material.

\begin{figure}[t]
\centering
\includegraphics[width=\textwidth]{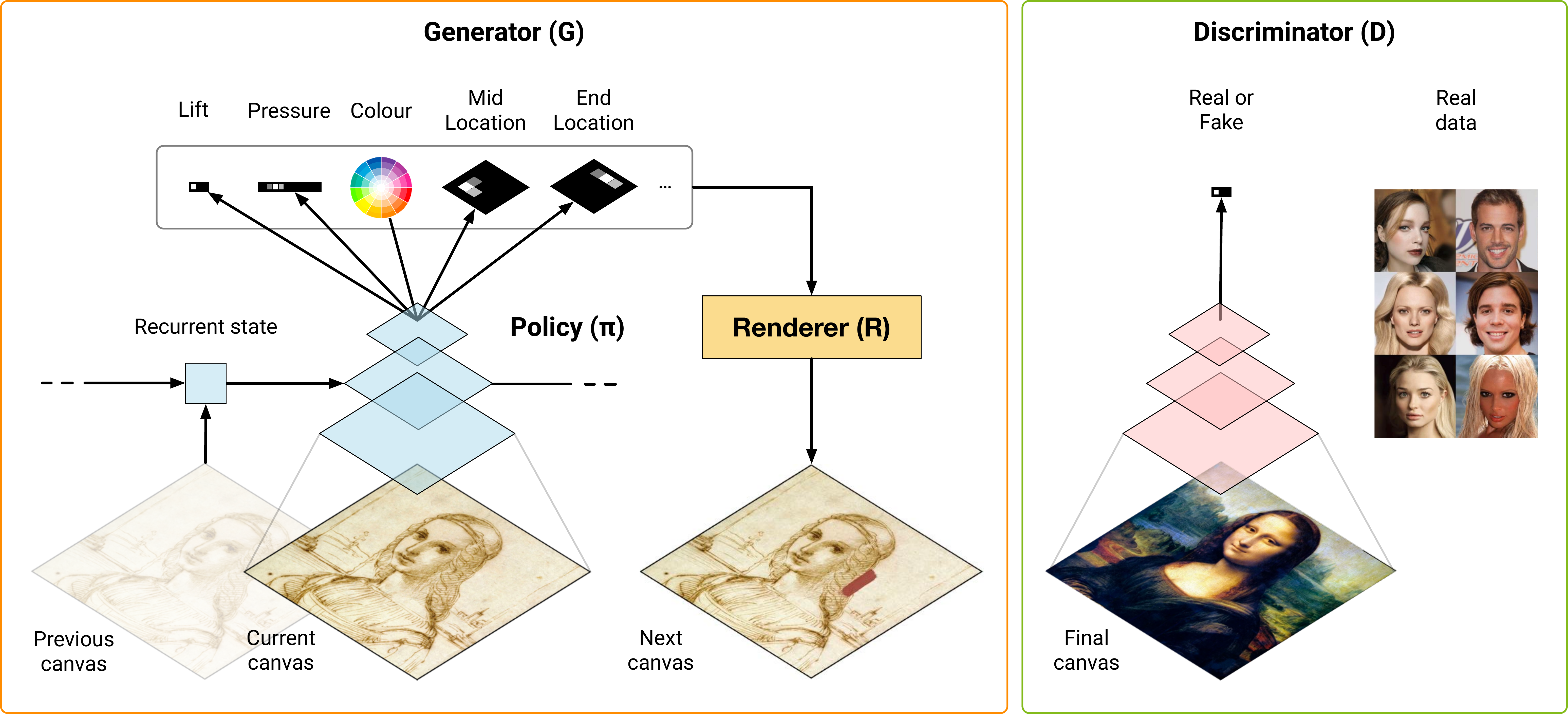}
\caption{\textbf{An architecture for \generativeagentsname.} There are two learnable components in the \oldmodelname architecture: the policy network or agent $ \pi $ and the discriminator network $ D $. The policy network takes as input a partially completed canvas and produces the parameters of the agent's action. This action is sent to the renderer $ \mathcal{R} $ as a command to update the canvas. The discriminator network takes as input the final canvas and attempts to classify it as real or fake. This network is trained using both images from the real dataset and rendered samples from the agent.}
\label{fig:architecture}
\end{figure}

\section{\newmodelname}
\label{sect:spiral_plusplus_improvements}

In this work, we build closely on the \oldmodelname architecture \citep{ganin2018spiral}, see \autoref{fig:architecture}. The goal of \oldmodelname is to train a policy $ \pi $ that controls a black-box rendering simulator $ \mathcal{R} $ using adversarial learning \citep{goodfellow2014generative}. At every time step, $ \pi $ observes the current state of the drawing and produces a rendering command $ a_t $ used by $ \mathcal{R} $ to update the canvas. Each command specifies the appearance of a single brush stroke (\eg, its shape and colour, the amount of pressure applied to the brush and so on -- the action space is discussed in greater detail in \autoref{sect:compound_action_space}). In \oldmodelname, the learning signal for the policy comes in the form of the reward computed by a discriminator network $ D $ simultaneously optimized to tell apart examples from some target distribution of images $ p_d $ and final renders produced by the agent (denoted as $ p_g $). The outcome of the training procedure is a policy that generates a sequences of commands such that $ p_g $ is close to $ p_d $. Although the \oldmodelname framework was shown to perform well on a number of tasks, in its original form it is limited in terms of image fidelity and scalability. We improve these aspects of the framework by introducing the simple but effective modifications detailed in the following subsections. We refer to the resulting model as \newmodelname in order to distinguish it from prior work.

\inlinesubsection{Spectral Normalization}
\label{sect:spectral_norm}

\oldmodelname \citep{ganin2018spiral} followed the setup of \cite{gulrajani2017wgangp}, namely that of a WGAN-GP discriminator architecture. In \newmodelname, we instead revert to the standard GAN loss \citep{goodfellow2014generative}, but with spectral normalization as described in \cite{miyato2018spectral}. Spectral normalization controls the Lipschitz constant of the discriminator by literally setting the spectral norm of each of its layers to equal 1. The motivation for this is to stabilize the training of the discriminator. In practice we find that spectral normalization significantly improves the fidelity of the generated images, as ablated in \autoref{sect:disc_reg_ablation} in the appendix.

\inlinesubsection{Temporal Credit Assignment}
\label{sect:temporal_credit_assignment}

\oldmodelname's reward structure (namely that of rewarding the agent only on the last step of each episode) provides agents with the flexibility to learn non-greedy image generation policies, however in practice, the sparsity of the learning signal limits the length of the episodes for which reinforcement learning techniques yield positive results. \cite{ganin2018spiral} only report results on episodes of length 20. It would be desirable to maintain this flexibility to achieve non-greedy policies whilst relaxing the difficulty of the learning problem. In \newmodelname, instead of only providing the agent a reward at the end of the episode based on the final discriminator loss, we calculate the discriminator loss at every step based on the partially-drawn canvas, and at every timestep the agent gets the 1-step improvement in loss as its reward: $ r_t = D(\mathcal{R}(a_{1:t})) - D(\mathcal{R}(a_{1:t-1})) $\footnote{Note lack of $ \gamma $ before the first term, unlike in \cite{ng1999rewardtransformations}.}.
Similarly to \cite{ng1999rewardtransformations}, this reward redistribution leads to the same optimal policies if the reinforcement learning discount factor $ \gamma $ is set to equal 1. Intuitively, when $ \gamma = 1 $, all terms cancel out in the calculation of returns apart from the final step's reward. However we obtained best results with $ \gamma $ between 0 and 0.99, which in this reward redistribution scheme causes the agent to be more greedy by introducing some bias \citep{guez2018mctsnets}. \cite{huang2019learningtopaint} similarly use this technique, concurrent with this work. We provide an ablation of the effectiveness of temporal credit assignment in \autoref{sect:ca_ablation} in the appendix.

\inlinesubsection{Complement Discriminator}
\label{sect:complement_discriminator}

When in conditional training mode, the discriminator is used to train agents such that when given a target image $ \xxsub{target} $, they produce reconstructions $ \xx $ that are as similar as possible to the target. Note that conditional agents are trained differently than their unconditional counterparts. \cite{ganin2018spiral} achieve this by conditioning the discriminator on the target. Specifically, in each forward pass, the input to the discriminator is either: 1. a \textit{fake} image pair $ (\xxsub{target}, \xx) $, or 2. a \textit{real} image pair $ (\xxsub{target}, \xxsub{target}) $. Agents are encouraged to move in a direction that makes renderings $ \xx $ more similar to the targets $ \xxsub{target} $ in order to fool the discriminator. A potential downfall of this method is that the discriminator can in theory find a simple shortcut to achieve perfect performance: it can compare the two images in each input pair, and if they are not identical, pixel to pixel, then it will know that it is a fake image pair. Whilst this strategy allows the discriminator to achieve its training objective, it can lead to a sub-optimal loss surface for the agent.

\begin{wrapfigure}{r}{0.19\textwidth}
    \vspace{-5mm}
    \centering\colorbox{black}{\begin{subfigure}[b]{0.09\textwidth}
        \adjincludegraphics[width=\textwidth,trim={0 0 {.76\width} 0},clip]{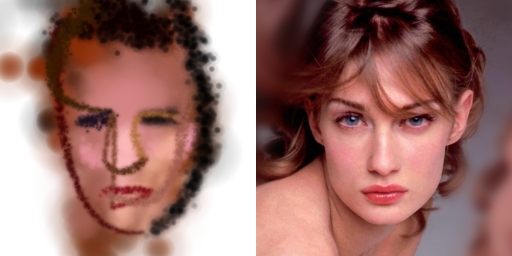}
    \end{subfigure}
    \begin{subfigure}[b]{0.09\textwidth}
        \adjincludegraphics[width=\textwidth,trim={{.76\width} 0 0 0},clip]{im/rtm/13fully_conditioned_pair}
    \end{subfigure}}
    \caption{Fake input \smash{$ (\xxsub{target}^{1-m}, \xx^m) $}}
    \label{fig:complement_discriminator_inputs}
\end{wrapfigure}
To overcome this we use what we call a \textit{complement discriminator}. In each forward pass, a binary image mask $ \mm $ is sampled. We mask the rendered image by $ \mm $ to obtain $ \xx^m = \mm \cdot \xx $, and mask the target both by $ \mm $ to obtain $ \xxsub{target}^m = \mm \cdot \xxsub{target} $ and by the complement of $ \mm $ to obtain \smash{$ \xxsub{target}^{1-m} = (1 - \mm) \cdot \xxsub{target} $}. We then define the fake input to the discriminator to be \smash{$ (\xxsub{target}^{1-m}, \xx^m) $} and the real input to the discriminator to be \smash{$ (\xxsub{target}^{1-m}, \xxsub{target}^m ) $}. In our experiments we mask the left/right half (see \eg \autoref{fig:complement_discriminator_inputs}) or top/bottom half of the image, however random masks are likely to work just as well \citep{pathak2016inpainting}. This trick makes it impossible for the discriminator to detect the real pair by simply comparing pixels, therefore creating a more suitable loss surface for the agent. We ablate the effect of the complement discriminator in \autoref{sect:complement_ablation} in the appendix.

\inlinesubsection{Population discriminator}
\label{sect:population_of_generators}

For best results we dynamically tune learning rate and entropy cost hyper-parameters using population based training (PBT, \citealp{jaderberg2017population}). However, unlike \cite{ganin2018spiral} which also employs this technique, we typically train with a periodically-evolving population of 10 \generativeagentsname sharing a \textit{single} discriminator. The effect of this modification is three-fold. Since the discriminator receives fake samples from all the policy learners, this means that it gets updated more frequently than individual generators allowing us to remove the replay buffer from the original \oldmodelname distributed setup. PBT also automatically resurrects individual generators should they collapse during training. Additionally each generator can now specialize in a subset of the modes of the full distribution, such that the population collectively covers the full distribution. Put another way, this technique allows us to increase the relative representational capacity of the generators as compared to the discriminator. We show in the experiments that this setup sometimes leads to the different agents in a single population learning different painting styles. Note, however, that we do not employ any explicit techniques to encourage or guarantee population diversity. 

\section{Experimental Results}
\label{sect:results}

Owing to the diverse range of settings under which the \generativeagentsname framework can be examined (with different action sets, rendering environments, brush types, episode lengths, agent and discriminator hyper-parameters, and so on), we first provide highlights of our main findings in \autoref{sect:results}. In figures, we show selected agents to illustrate phenomena that are \textit{possible} to observe with the framework. This is not to imply that the settings those agents were trained with are necessary or sufficient for the observed results. In the appendix we provide a systematic analysis of the major components of \newmodelname via controlled ablations.

\label{sect:experiments}

\label{sect:results_celeba}

\begin{figure}[t]
\centering
(a)\hfill\raisebox{-.5\height}{\includegraphics[width=0.96\textwidth]{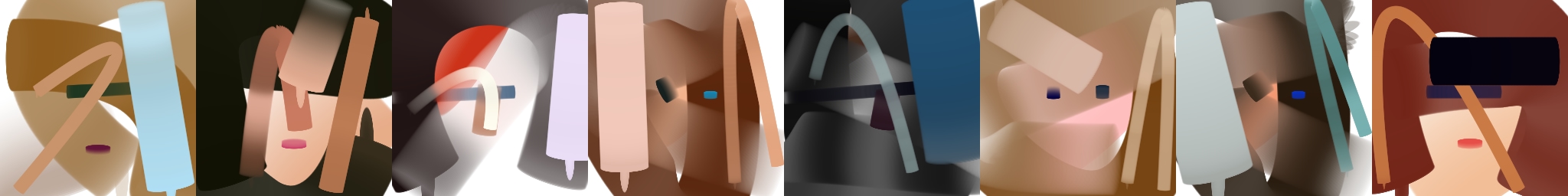}} \\
(b)\hfill\raisebox{-.5\height}{\includegraphics[width=0.96\textwidth]{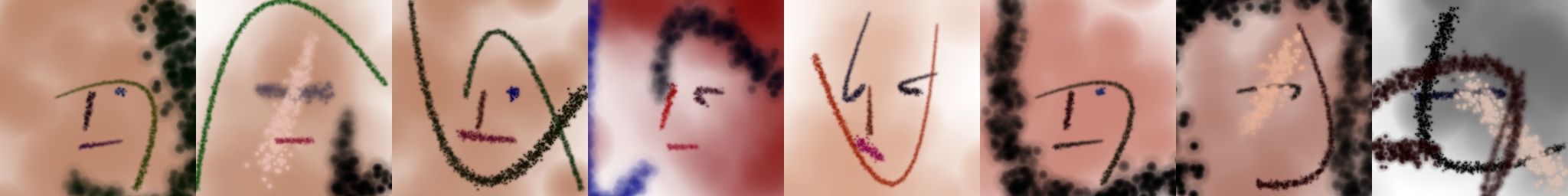}} \\
(c)\hfill\raisebox{-.5\height}{\includegraphics[width=0.96\textwidth]{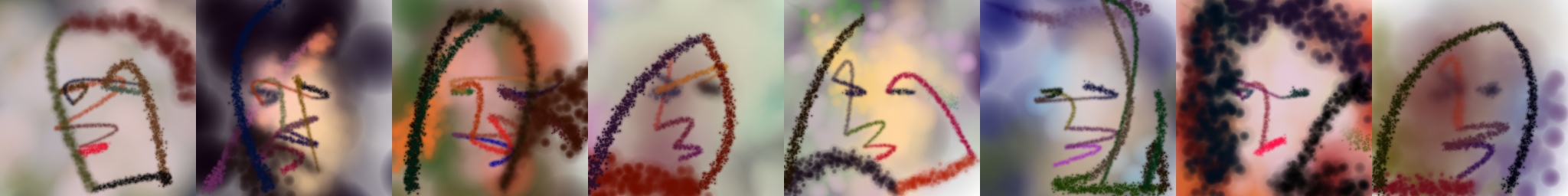}} \\
(d)\hfill\raisebox{-.5\height}{\includegraphics[width=0.96\textwidth]{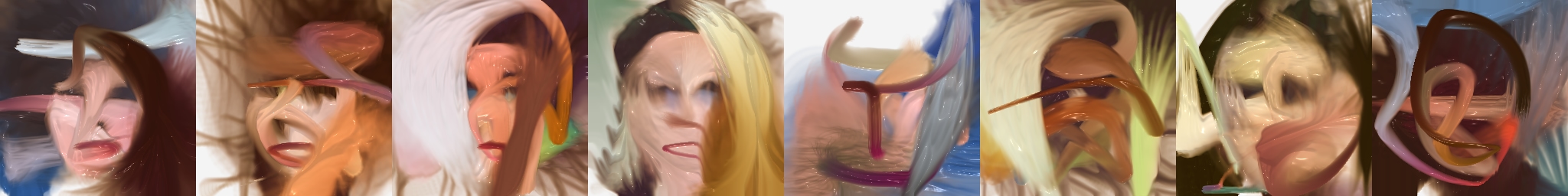}} \\
(e)\hfill\raisebox{-.5\height}{\includegraphics[width=0.96\textwidth]{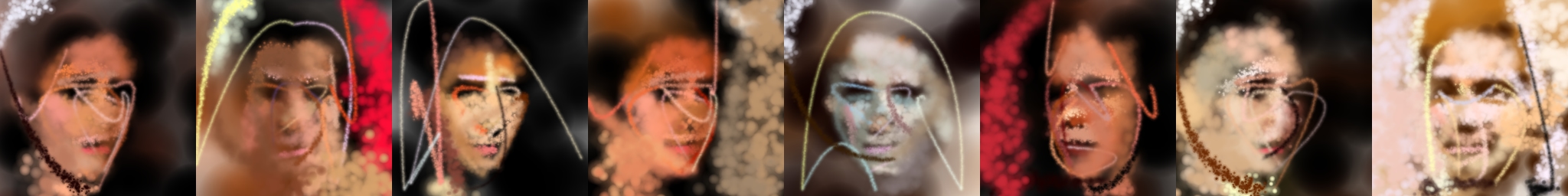}} \\
(f)\hfill\raisebox{-.5\height}{\includegraphics[width=0.96\textwidth]{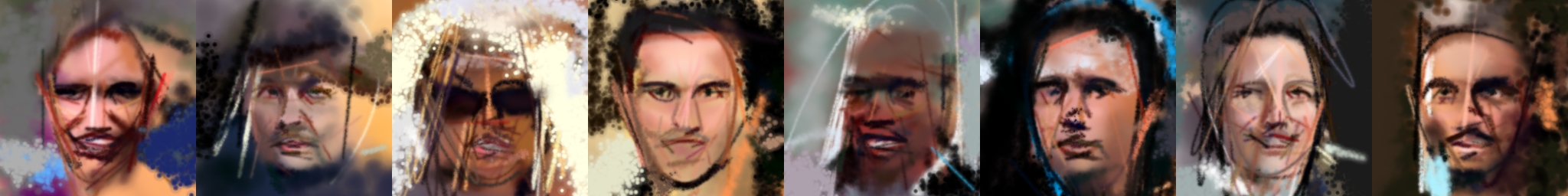}} \\
(g)\hfill\raisebox{-.5\height}{\includegraphics[width=0.96\textwidth]{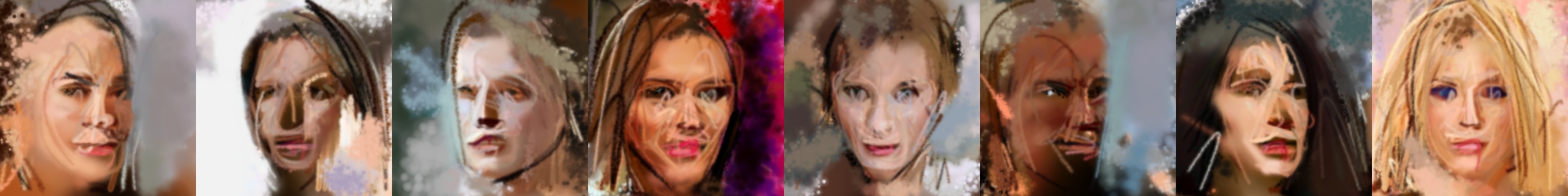}} \\
(h)\hfill\raisebox{-.5\height}{\includegraphics[width=0.96\textwidth]{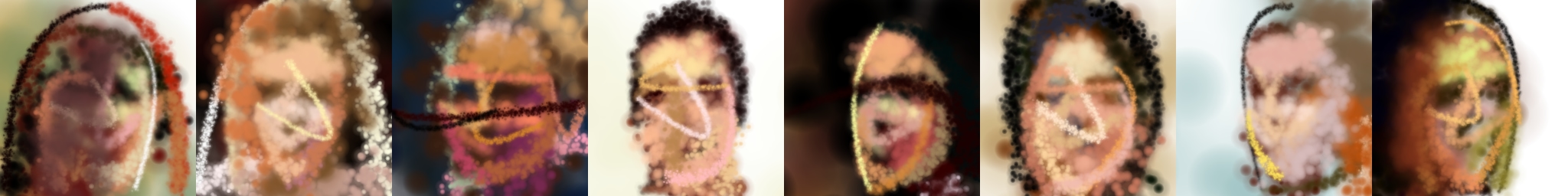}} \\
\caption{\textbf{Artificial portraiture with \generativeagentsname.} Random unconditional samples generated by agents trained on CelebA-HQ. Separate agents with varying hyper-parameters were trained to generate samples (a, b, c) in 17 steps with various brushes and action spaces, (d) in 19 steps using an oil paint simulator \citep{li2017fluid}, (e) in 31 steps, (f) in 400 steps, (g) in 1000 steps. As a baseline, (h) shows unconditional samples generated in 19 steps by our best reimplementation of \oldmodelname (\citealp{ganin2018spiral}, WGAN-GP + single discriminator per population + hyperparameter tweaks). Our improved method scales with episode length from relatively abstract to approach realistic-looking results. See \url{https://learning-to-paint.github.io} for videos.}
\label{fig:gen_episode_length_abstraction}
\end{figure}

We show highlights of our experiments on Celeba-HQ \citep{karras2017progressive} in \autoref{fig:gen_episode_length_abstraction}. Amongst all the framework's settings, the one we observed to have the most profound impact on the agents' behaviour was the number of brush strokes (steps) they were allowed in each episode to generate an image. Agents that were constrained with short episodes learned qualitatively different policies than those that could afford numerous interactions, producing images with a degree of visual abstraction, not unlike how humans do when similarly constrained \citep{selim2018spiderman, fan2019pragmatic}. For this reason we structure the results into two sections: short episodes in \autoref{sect:results_celeba_short} and long episodes in \autoref{sect:results_celeba_long}.

\inlinesubsection{Short episodes}
\label{sect:results_celeba_short}

We encourage the reader to take a few moments to inspect the samples in \autoref{fig:gen_episode_length_abstraction}(a-c). Each row was generated by a different agent, differing in the settings of their environments (\eg brush type, action space, episode length).

It was surprising for us to see the aesthetically pleasing way in which the agent draws faces: \eg using a large circle to delineate the outline of the face, dots for each of the eyes, a line for the nose and a line for the mouth. Note that the agent has never been exposed to human \textit{drawings} of human faces, but only to realistic \textit{photographs} of human faces. Also note that the agents choose to use bright colours and thin strokes to depict salient elements of faces despite no element of the framework encouraging such behaviour. In all cases, the architecture of the agent is constant, and it is the variation in the characteristics of the agent's environment that creates the diversity of styles. These results serve as an existence proof for the conjecture that some aspects of human drawings of faces can emerge automatically from a learning framework as simple as that of \generativeagentsname (namely agents equipped with brushes working against a discriminator), without the need for supervision, imitation or social cues.

\begin{figure}[t]
\centering

\begin{subfigure}[b]{0.32\textwidth}
    \adjincludegraphics[width=\textwidth, trim={0 {.5\height} 0 0}, clip]{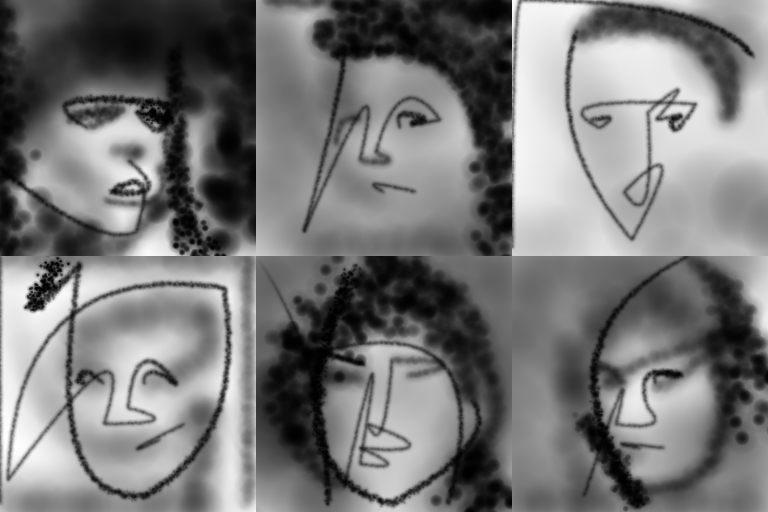}
    \caption{gray with gray (diverse)}
\end{subfigure}
\hfill
\begin{subfigure}[b]{0.32\textwidth}
    \adjincludegraphics[width=\textwidth, trim={0 {.5\height} 0 0}, clip]{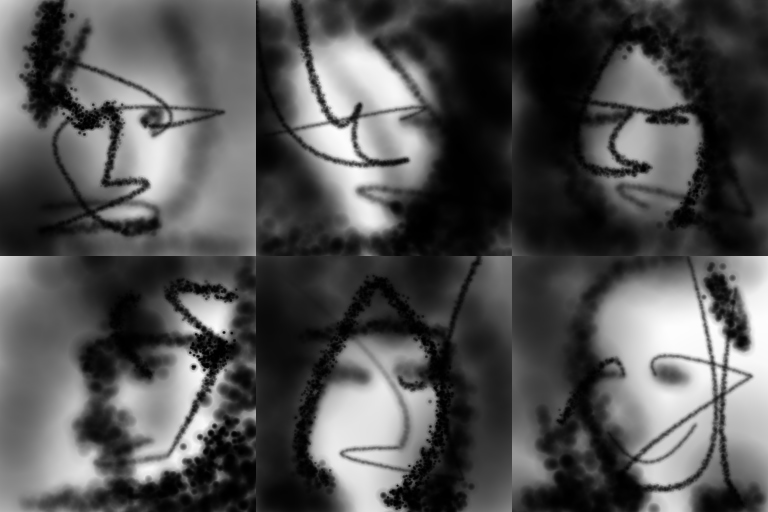}
    \caption{color with gray (mode-collapse)}
\end{subfigure}
\hfill
\begin{subfigure}[b]{0.32\textwidth}
    \adjincludegraphics[width=\textwidth, trim={0 {.49807\height} 0 0}, clip]{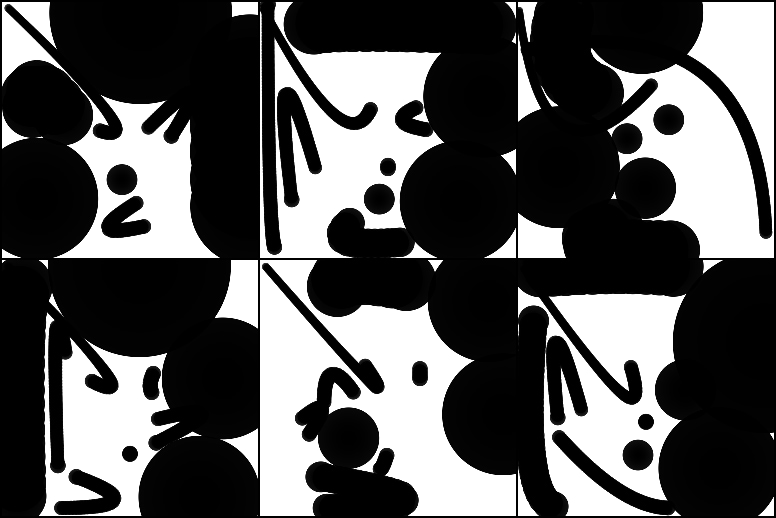}  %
    \caption{gray with black (mode-collapse)}
\end{subfigure}

\caption{\textbf{Tasked with the impossible, these agents make do.} Samples are selected from three 20 step models trained on CelebA-HQ with modified environments. As a baseline, (a) was tasked with generating grayscale photos using a black brush with variable opacity. (b) was tasked with generating color photos using the same variable opacity black brush. (c) was tasked with generating grayscale photos using an opaque black pen. Models (b) and (c) often manage to draw recognizable faces despite the huge gap between what they \textit{can} and \textit{should} produce, however both experience severe mode collapse: each agent in these populations generates minor variations on a single image rather than a full distribution of images.}
\label{fig:color_ablation}
\end{figure}

Unlike in most modern GANs which directly output pixels, in this setting there is a large discrepancy between what the generator \textit{can} produce, and what it \textit{should} produce. The brushes are too constrained and there is simply not enough time in the episodes for the generator to be able to produce a completely photo-realistic image. And in our experiments the discriminators can always distinguish between generated and real images with high confidence. Nevertheless we observe that when sufficiently regularised, discriminators can provide sufficient signal for meaningful learning to take place, suggesting that they are still capable of ranking generated images in a useful way. We explore this further in \autoref{fig:color_ablation}, by training agents on color photos but only with various grayscale brushes.

\begin{figure}[t]
\centering
\adjincludegraphics[width=\textwidth, trim={{.083\width} 0 0 0}, clip]{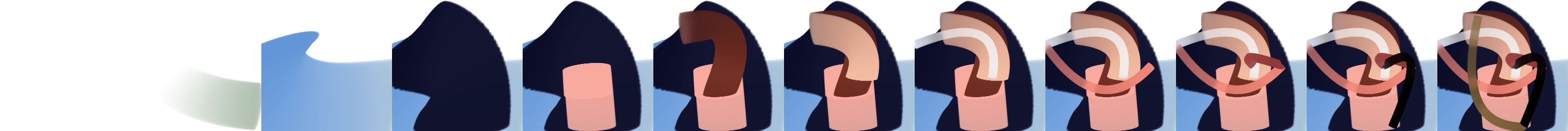}
\par\medskip
\adjincludegraphics[width=\textwidth, trim={{.083\width} 0 0 0}, clip]{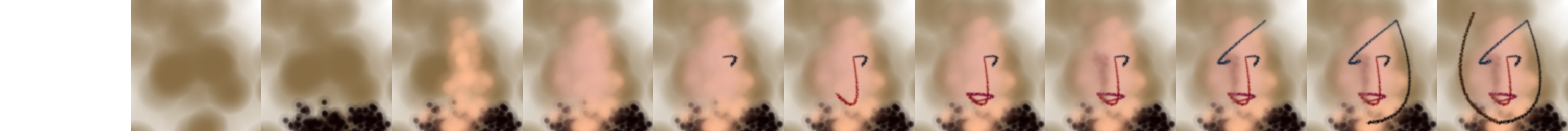}
\caption{\textbf{10 step canvas sequences show precise control.} Agents interact with the canvas to produce the final image, manipulating the location, colour and thickness of the brush with precision. See \url{https://learning-to-paint.github.io} for videos. \label{fig:gen_20step_filmstrips}}
\vspace{-3mm}
\end{figure}

It is informative to examine how agents interact with the simulated canvas to produce these images. In \autoref{fig:gen_20step_filmstrips} we show a number of episodes as sequences. We see that agents can learn to manipulate the location, colour and thickness of the brush with precision, constructing the final images stroke by stroke. It is worth noting that due to RL's objective of maximizing potentially delayed rewards, agent policies often deviate from being purely greedy, and they often take actions that appear to reduce the quality of the image, especially early in the episode, only for it to be revealed to have been important for the final drawing. We revisit this point in \autoref{sect:results_celeba_long} and \autoref{fig:rec_distances} in the appendix.

\begin{figure}[t]
\centering
\includegraphics[width=\textwidth]{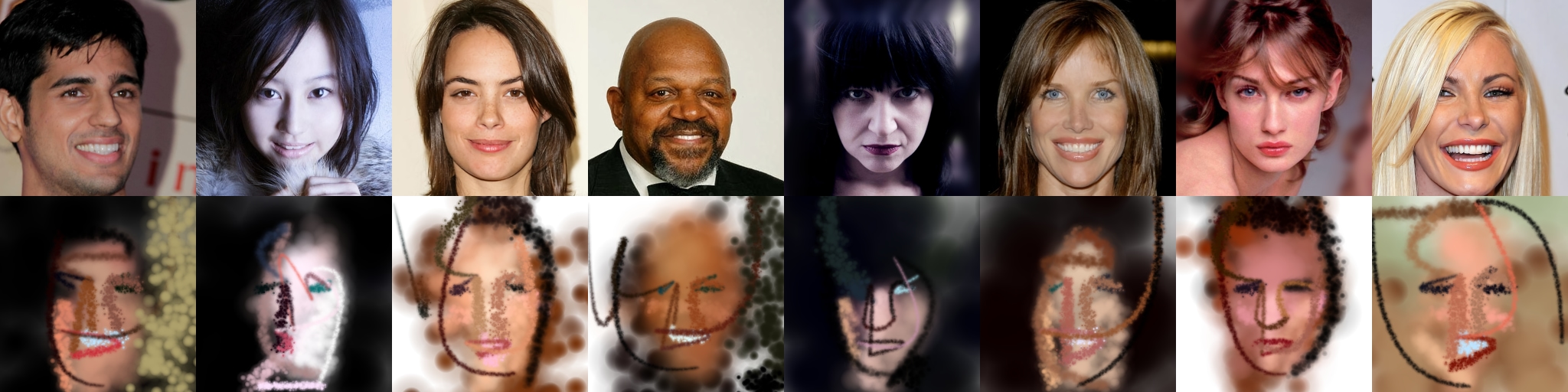}
\caption{\textbf{Conditional image generation.} The agent is generally able to match the higher level statistics of the targets (top), and even appears to capture the faces' smiles.}
\label{fig:rec_20steps_samplesheet}
\end{figure}

As described in \autoref{sect:population_of_generators}, we use population based training (PBT, \citealp{jaderberg2017population}). In \autoref{fig:gen_20steps_multi} (and \autoref{fig:gen_200steps_multi} in the appendix) we show samples from three different agents that were trained as part of the same population. We see that the different generators each specialize in a subset of the modes of the full distribution, each producing images with a perceptibly different style. Note that in all three cases the architecture of the agents, their action spaces, and the settings of their environments were identical, and they differ only in their weights and evolved learning rate and entropy cost. Finally, we show conditional samples in \autoref{fig:rec_20steps_samplesheet}. The agent is able to match the higher level statistics of the target images, and even appears to capture the faces' smiles. In \autoref{fig:rec_fixed_target} in the appendix we visualise the agent's stochasticity by producing multiple samples for the same target image.

\begin{figure}[t]
\centering

\begin{subfigure}[b]{0.32\textwidth}
    \adjincludegraphics[width=\textwidth, trim={0 {.5\height} {.25\width} 0}, clip]{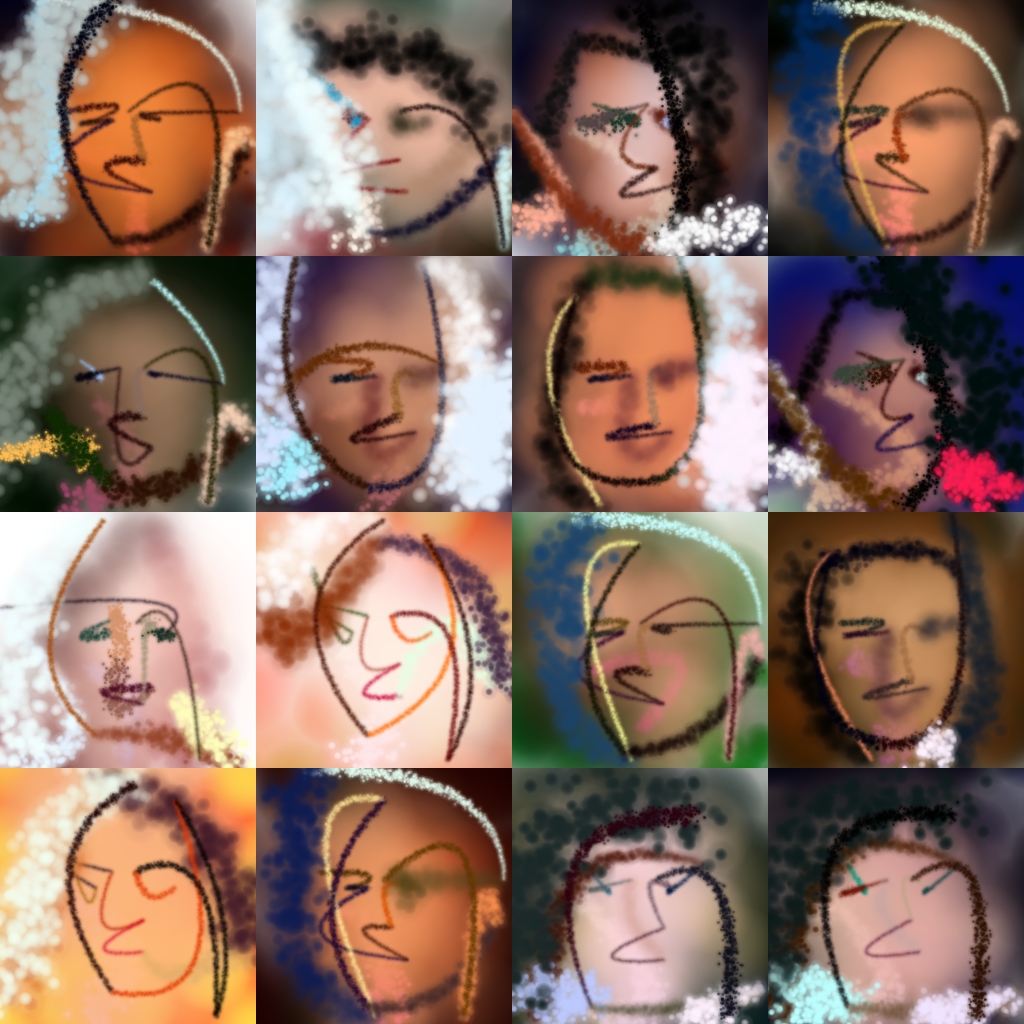}
\end{subfigure}
\hfill
\begin{subfigure}[b]{0.32\textwidth}
    \adjincludegraphics[width=\textwidth, trim={0 {.5\height} {.25\width} 0}, clip]{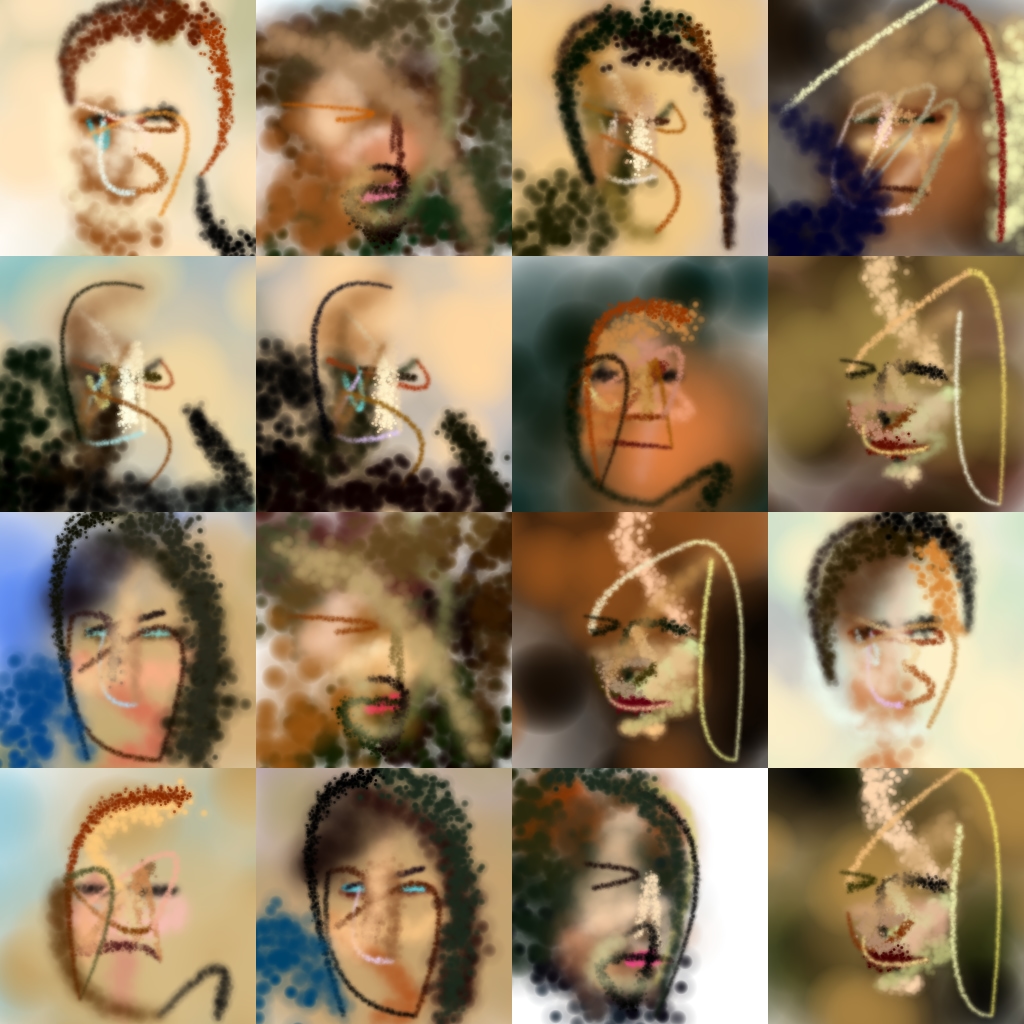}
\end{subfigure}
\hfill
\begin{subfigure}[b]{0.32\textwidth}
    \adjincludegraphics[width=\textwidth, trim={0 {.5\height} {.25\width} 0}, clip]{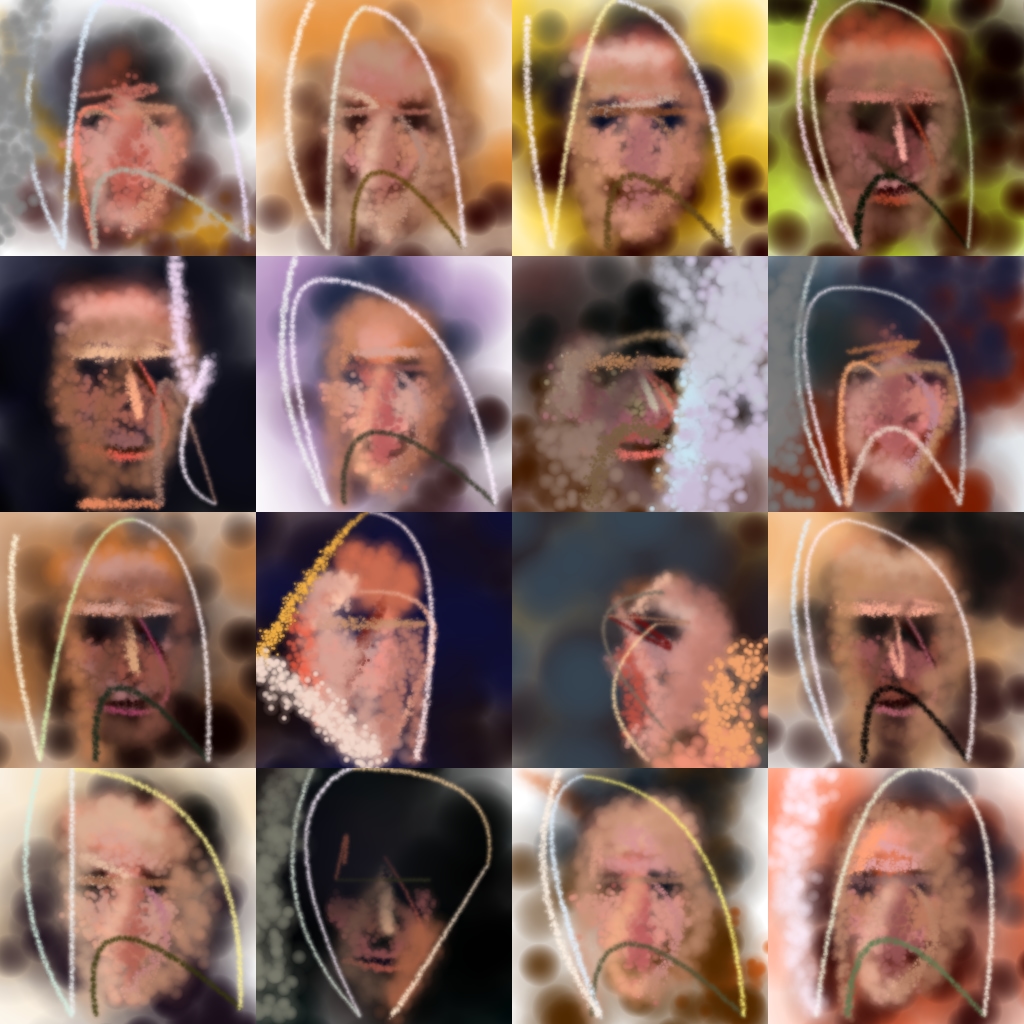}
\end{subfigure}

\caption{\textbf{Within-population and within-agent diversity (short episodes).} We show three sets of samples for three different agents in the same population, showing both the diversity of samples for each agent, and the diversity of samples across agents in the same population. The first agent attempts more figurative line drawings, whilst the agent on the right uses more realistic shading. We do occasionally observe `mode collapse' where certain samples repeat themselves, for instance in the middle agent though there is still slight variation in the way each image is drawn.
\label{fig:gen_20steps_multi}}
\end{figure}

\inlinesubsection{Long episodes}
\label{sect:results_celeba_long}

We found temporal credit assignment (TCA, \autoref{sect:temporal_credit_assignment}) to be crucial for training agents on long episodes. Without TCA, agents only learn to perform meaningful actions in the last 15 to 60 steps or so of the episode, regardless of the episode length. However, with TCA, agents make full use of episodes of up to 1000 steps, leading in turn to qualitatively different generation policies. We train with discounts of 0.9 or 0.99 and unroll lengths of 20-50 in the experiments that follow. See \autoref{fig:gen_episode_length_abstraction}(f,g) for samples (as well as \autoref{fig:gen_200steps_multi} and \autoref{fig:improvements} in the appendix).

In \autoref{fig:gen_1000step_filmstrips} we show a visualisation of how images are constructed by TCA agents. It can be seen that agents are successful in making good use of hundreds of timesteps, controlling the brush with precision to form comparatively realistic images. Due to the smaller-than-1 discount factor, agents are incentivised to have somewhat believable intermediate canvases, leading to the formation of the most critical elements of faces (\eg eyes, nose and mouth) even in the earliest steps of the episode. In unconditional generation, much of the form of the final is determined by the randomness of the first few strokes on the canvas, with the agent doing its best towards the end of the episode to complete the result. In this light, the agent can be seen to be similar in spirit to auto-regressive models such as Pixel-RNNs and Pixel-CNNs \citep{oord2016pixel}, but in stroke-space as opposed to pixel-space.

\begin{figure}[t]
\centering
\includegraphics[width=\textwidth]{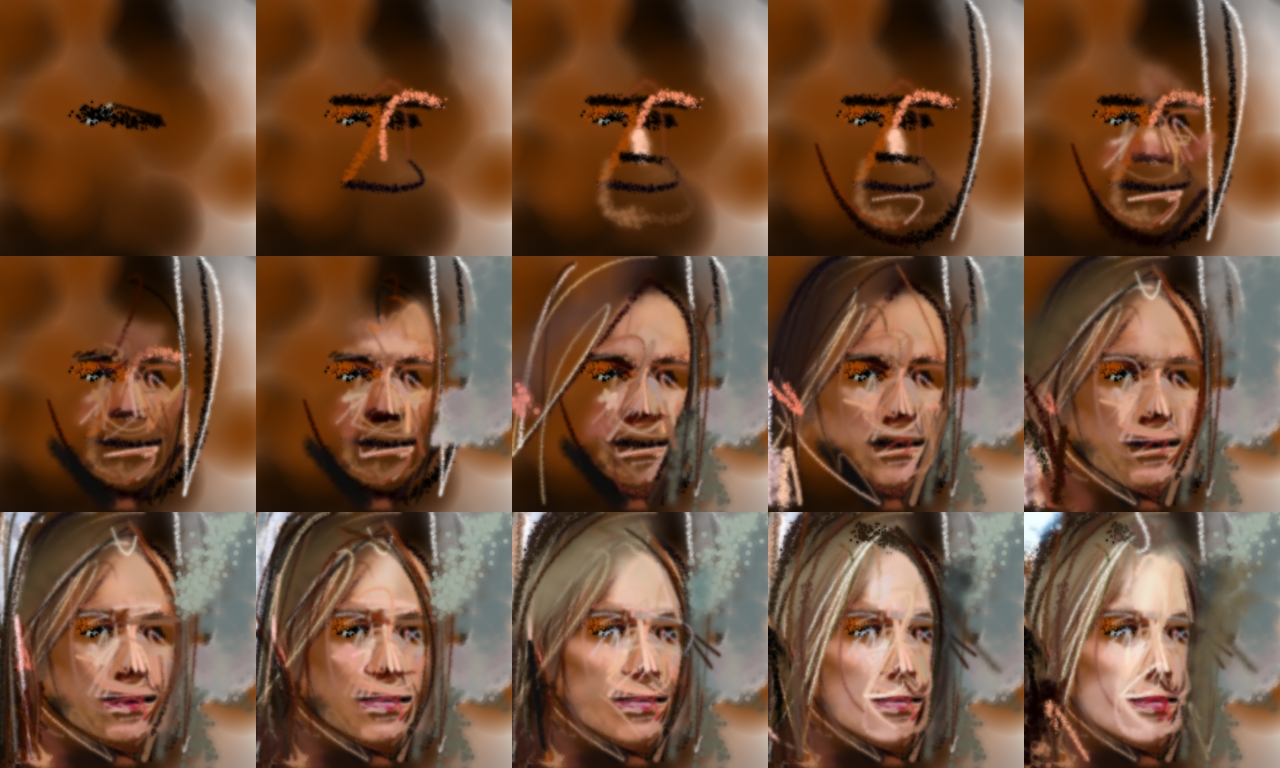}
\par\medskip
\includegraphics[width=\textwidth]{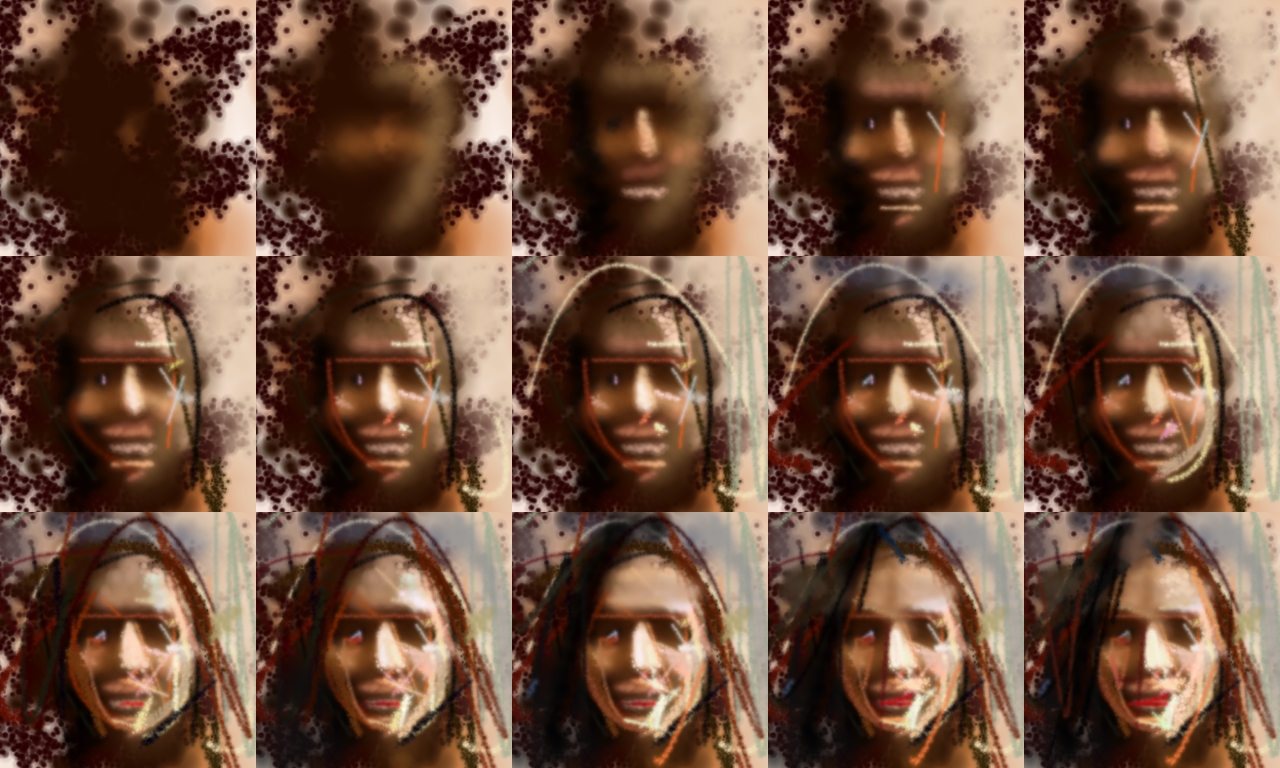}
\caption{\textbf{Longer sequences demonstrate higher levels of realism.} \newmodelname agents make good use of hundreds of timesteps, controlling the brush with precision. Here we show
logarithmically spaced frames from a 1000 and 500 step agent. See \url{https://learning-to-paint.github.io} for videos. \label{fig:gen_1000step_filmstrips}}
\end{figure}

\inlinesubsection{Other datasets and environments}
\label{sect:other_results}

We conduct a set of additional experiments to assess the generality of the proposed approach. \autoref{sect:results_imagenet} presents results for several ImageNet \citep{russakovsky2015imagenet} classes. In \autoref{sect:results_omniglot}, we discuss how a slight modification to the painting environment can lead to meaningful parses of the Omniglot \citep{lake2015human} characters. Finally, \autoref{sect:comparison_with_model_based} introduces a new oil paint simulator and shows that our agent is able to learn to control it despite the added complexity of the setting. 

\section{Related Work}

Non-photorealistic rendering emerged as an area of digital art that focuses on enabling a wide variety of artistic styles (in contrast to standard computer graphics which has a focus on photorealism). Stroke based rendering is a particular approach in non-photorealistic rendering that creates images by placing discrete elements such as paint strokes or stipples (see \citealp{hertzmann2003survey} for a comprehensive review). A common goal in stroke based rendering is to make the painting look like some other image, whilst limiting the total number of strokes \citep{hertzmann2003survey}. Techniques from computer vision have in some cases been used to control the positions of the strokes \citep{zeng2009image}. One attractive property of stroke based rendering is that its interpretations of images can, in theory, be executed in the physical world using real robots \citep{kotani19teaching, el2019mobile}. Closely connected are notions of `low-complexity' art \citep{schmidhuber1997low}, which describe an algorithmic theory of beauty and aesthetics based on the principles of information theory. It postulates that amongst several comparable images, the most pleasing one is the one with the shortest description. In almost all cases, the aforementioned techniques operate by optimising, either explicitly or implicitly, a fixed objective function without any learnable parameters. New large-scale sources of data have made it possible to model human behaviour more directly in simple drawing domains \citep{ha2017neural, zhou2018learning, li2019photosketching}, potentially making it possible to capture our internal object function in a data-driven manner.

Superficially, \oldmodelname \citep{ganin2018spiral} and \newmodelname can be seen as a technique for stroke based, non-photorealistic rendering. Similarly to some modern stroke based rendering techniques, the positions of strokes are determined by a learned system (namely, the agent). Unlike traditional methods however, the objective function that determines the goodness of the output image is learned unsupervised via the adversarial objective. This adversarial objective allows us to train without access to ground-truth strokes, in contrast to \eg \cite{ha2017neural, zhou2018learning, li2019photosketching}, enabling the method to be applied to domains where labels are prohibitively expensive to collect, and for it to discover surprising and unexpected styles.

There are a number of works that use constructions similar to \oldmodelname to tune the parameters of non-differentiable simulators \citep{louppe2017adversarial, ruiz2018learning, kar2019meta}. \cite{frans2018unsupervised, nakano2019neural, zheng2018strokenet, huang2019learningtopaint} achieve remarkable learning speed by relying on back-propagation through a learned model of the renderer, however they rely on being able to train an accurate model of the environment in the first instance. We further elaborate on this difference in \autoref{sect:comparison_with_model_based}.
On the applications side, \cite{el2018keep, lazaro2018beyond, agrawal2018generating} build systems that generate outcomes iteratively, conditioned on ongoing linguistic input or feedback. \cite{ellis2018learning, liu2018learning} use inference techniques to convert simple scenes into programs expressing abstract regularities. Technical advances such as these (for instance the use of neural simulators and model-based RL) are largely orthogonal to what we introduce in this paper, and are likely to improve the performance of \newmodelname further when used in combination. Nonetheless we note that to the best of our knowledge, very few models have the capability to paint images without conditioning on a target image, and even fewer do so at a similar level of abstraction, and lead to as many surprising, emergent styles as those obtained with \newmodelname.

\section{Discussion and Conclusions}

\begin{figure}[h]
\centering
\includegraphics[width=\textwidth]{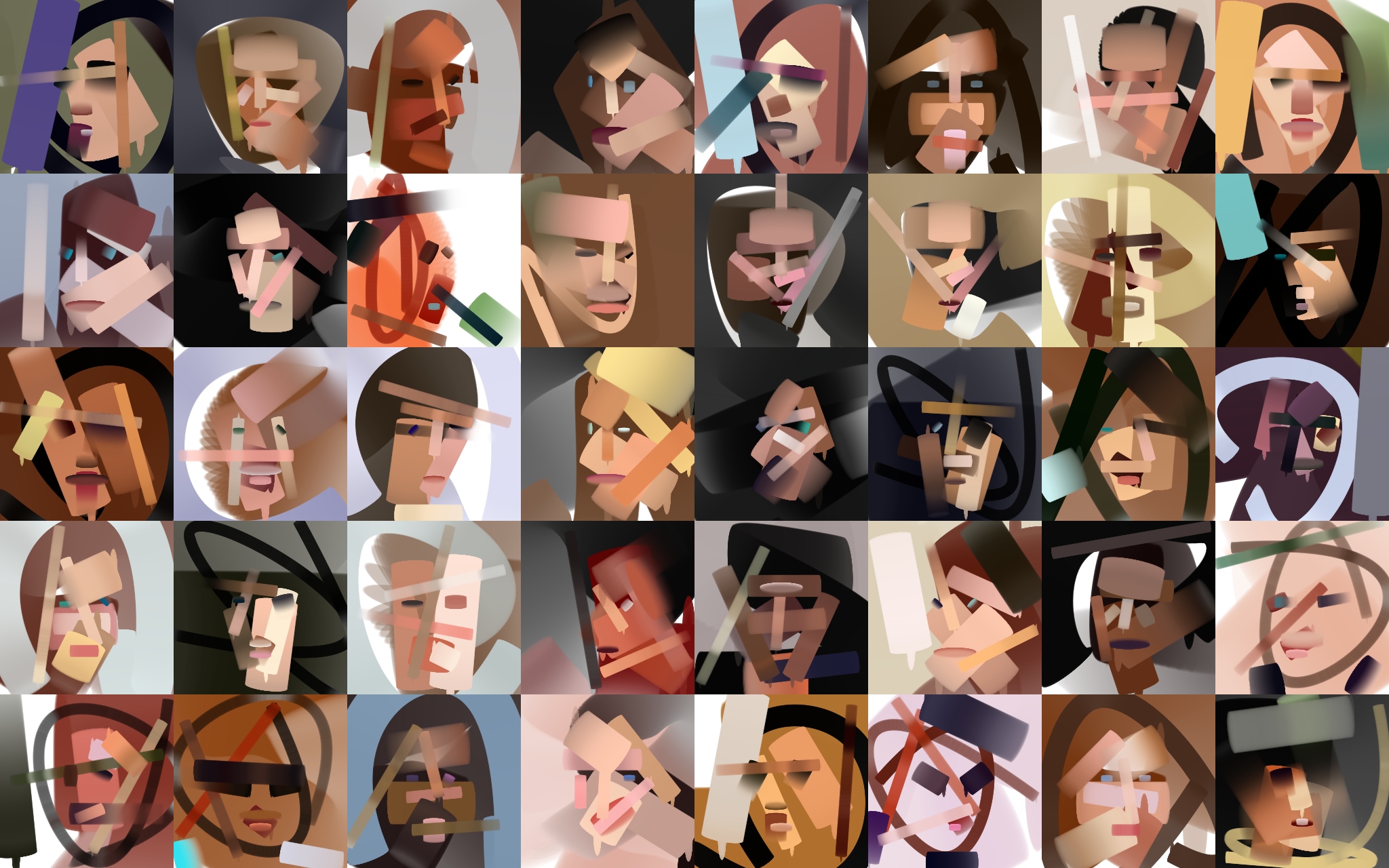}
\caption{Selected samples from a single population of agents. The agents find a great variety of ways of working within the constraints of the brush they were given to evoke faces in just 32 steps.}
\label{fig:creative}
\end{figure}

We showed that when sufficiently constrained, \generativeagentsname can learn to produce images with a degree of visual abstraction, despite having never been shown human drawings. To the best of our knowledge, this is the first time that an unsupervised artificial system has discovered visual abstractions of this kind, abstractions that resemble those made by children and novice illustrators. We stress that our aim has been to show that certain results are at least possible to achieve with the \generativeagentsname framework, not that the specifics of our setting are necessarily the best way of achieving those results. It is important to acknowledge that our results are mostly qualitative, in part due to the subjective nature of the phenomena being studied (\eg improved realism and abstraction as compared to \oldmodelname). Nevertheless an important line of work will be to quantify these effects via controlled human studies, or even by exposing the generated samples to the markets \citep{cohn2018art}! 

The abstraction results arise due to the constraints imposed on the generator by the renderer and the episode's finite horizon. The generator cannot manipulate pixels directly in the way that generative models typically do; instead it has to obey the physics of the rendering simulation. In other words, the generator is \textit{embodied}, albeit in a simulation. The discriminator must also be well regularized for this result. It is worth noting that the framework does utilize structure to obtain its results, but this structure is built into the environment, not the agent itself. This stands in contrast with efforts that achieve abstraction by structuring the model or agent, and provides a potentially useful data point for recent debates around the utility of structure (see \eg \citealp{sutton2019bitter, welling2019do}). 

We also showed that such \generativeagentsname can produce outputs with considerable realism and complexity when given enough time with the environment. This is a considerable step up from the original \oldmodelname, providing encouraging evidence that the approach can be scaled to other sequence generation tasks where expert sequences are difficult to obtain for imitation, for instance in music generation, program synthesis, chemical synthesis and protein folding.

Finally, we note the framework's potential for creative and artistic applications. Machine learning techniques such as GANs \citep{goodfellow2014generative} and neural style transfer \citep{gatys2015neural} are increasingly being used for creative investigation in the artistic community. Generative agents have an uncommon ability to generate novel aesthetic styles with no human input except for a choice of brush (see \eg \autoref{fig:creative}). The extent to which any of these systems can be considered \textit{creative} in and of themselves is open for discussion \citep{hertzmann2018can}, however we look forward to seeing whether the proposed framework can provide new points for consideration in that debate.

\paragraph{Acknowledgements}

We would like to thank Aishwarya Agrawal, Nicolas Heess, Simon Kohl, Adam Kosiorek and David Ha for insightful discussions and comments on this manuscript.

\bibliographystyle{hhumannat}
\bibliography{bibliography}

\begin{thebibliography}{}

\bibitem[\protect\astroncite{Agrawal et~al.}{2018}]{agrawal2018generating}
Agrawal, A., M.~Malinowski, F.~Hill, S.~M.~A. Eslami, O.~Vinyals, and
  T.~Kulkarni\leavevmode\nopagebreak\newline 2018.
\newblock Generating diverse programs with instruction conditioned reinforced
  adversarial learning.
\newblock \href{https://arxiv.org/abs/1812.00898}{{\ttfamily arXiv:1812.00898
  [cs.LG]}}.

\bibitem[\protect\astroncite{Brock et~al.}{2019}]{brock2018biggan}
Brock, A., J.~Donahue, and K.~Simonyan\leavevmode\nopagebreak\newline 2019.
\newblock Large scale {GAN} training for high fidelity natural image synthesis.
\newblock In {\em International Conference on Learning Representations}.

\bibitem[\protect\astroncite{Clark and Chalmers}{1998}]{clark1998extended}
Clark, A. and D.~Chalmers\leavevmode\nopagebreak\newline 1998.
\newblock The extended mind.
\newblock {\em analysis}, 58(1):7--19.

\bibitem[\protect\astroncite{Cohn}{2018}]{cohn2018art}
Cohn, G.\leavevmode\nopagebreak\newline 2018.
\newblock {AI} art at christie's sells for \$432,500.
\newblock
  \url{https://www.nytimes.com/2018/10/25/arts/design/ai-art-sold-christies.html}.

\bibitem[\protect\astroncite{El~Helou et~al.}{2019}]{el2019mobile}
El~Helou, M., S.~Mandt, A.~Krause, and
  P.~Beardsley\leavevmode\nopagebreak\newline 2019.
\newblock Mobile robotic painting of texture.
\newblock {\em 2019 International Conference on Robotics and Automation
  (ICRA)}, Pp.~ 640--647.

\bibitem[\protect\astroncite{El-Nouby et~al.}{2018}]{el2018keep}
El-Nouby, A., S.~Sharma, H.~Schulz, D.~Hjelm, L.~El~Asri, S.~E. Kahou,
  Y.~Bengio, and G.~W. Taylor\leavevmode\nopagebreak\newline 2018.
\newblock Tell, draw, and repeat: Generating and modifying images based on
  continual linguistic instruction.
\newblock \href{https://arxiv.org/abs/1811.09845}{{\ttfamily arXiv:1811.09845
  [cs.CV]}}.

\bibitem[\protect\astroncite{Ellis et~al.}{2018}]{ellis2018learning}
Ellis, K., D.~Ritchie, A.~Solar-Lezama, and
  J.~Tenenbaum\leavevmode\nopagebreak\newline 2018.
\newblock Learning to infer graphics programs from hand-drawn images.
\newblock In {\em Advances in Neural Information Processing Systems}, Pp.~
  6059--6068.

\bibitem[\protect\astroncite{Eslami et~al.}{2018}]{eslami2018neural}
Eslami, S. M.~A., D.~J. Rezende, F.~Besse, F.~Viola, A.~S. Morcos, M.~Garnelo,
  A.~Ruderman, A.~A. Rusu, I.~Danihelka, K.~Gregor, and
  {others}\leavevmode\nopagebreak\newline 2018.
\newblock Neural scene representation and rendering.
\newblock {\em Science}, 360(6394):1204--1210.

\bibitem[\protect\astroncite{Fan et~al.}{2019}]{fan2019pragmatic}
Fan, J.~E., R.~D. Hawkins, M.~Wu, and N.~D.
  Goodman\leavevmode\nopagebreak\newline 2019.
\newblock Pragmatic inference and visual abstraction enable contextual
  flexibility during visual communication.
\newblock {\em Computational Brain \& Behavior}.

\bibitem[\protect\astroncite{Fernando and {others}}{2004}]{fernando2004gpu}
Fernando, R. and {others}\leavevmode\nopagebreak\newline 2004.
\newblock {\em GPU gems: programming techniques, tips, and tricks for real-time
  graphics}, volume 590.
\newblock Addison-Wesley Reading.

\bibitem[\protect\astroncite{Frans and Cheng}{2018}]{frans2018unsupervised}
Frans, K. and C.-Y. Cheng\leavevmode\nopagebreak\newline 2018.
\newblock Unsupervised image to sequence translation with canvas-drawer
  networks.
\newblock \href{https://arxiv.org/abs/1809.08340}{{\ttfamily arXiv:1809.08340
  [cs.CV]}}.

\bibitem[\protect\astroncite{Ganin et~al.}{2018}]{ganin2018spiral}
Ganin, Y., T.~Kulkarni, I.~Babuschkin, S.~M.~A. Eslami, and
  O.~Vinyals\leavevmode\nopagebreak\newline 2018.
\newblock Synthesizing programs for images using reinforced adversarial
  learning.
\newblock In {\em Proceedings of the 35th International Conference on Machine
  Learning}.

\bibitem[\protect\astroncite{Gatys et~al.}{2016}]{gatys2015neural}
Gatys, L., A.~Ecker, and M.~Bethge\leavevmode\nopagebreak\newline 2016.
\newblock A neural algorithm of artistic style.
\newblock {\em Journal of Vision}, 16(12):326.

\bibitem[\protect\astroncite{Goodfellow
  et~al.}{2014}]{goodfellow2014generative}
Goodfellow, I., J.~Pouget-Abadie, M.~Mirza, B.~Xu, D.~Warde-Farley, S.~Ozair,
  A.~Courville, and Y.~Bengio\leavevmode\nopagebreak\newline 2014.
\newblock Generative adversarial nets.
\newblock In {\em Advances in neural information processing systems}, Pp.~
  2672--2680.

\bibitem[\protect\astroncite{Gregor et~al.}{2015}]{gregor2015draw}
Gregor, K., I.~Danihelka, A.~Graves, D.~J. Rezende, and
  D.~Wierstra\leavevmode\nopagebreak\newline 2015.
\newblock Draw: A recurrent neural network for image generation.
\newblock In {\em ICML}.

\bibitem[\protect\astroncite{Guez et~al.}{2018}]{guez2018mctsnets}
Guez, A., T.~Weber, I.~Antonoglou, K.~Simonyan, O.~Vinyals, D.~Wierstra,
  R.~Munos, and D.~Silver\leavevmode\nopagebreak\newline 2018.
\newblock Learning to search with {MCTS}nets.
\newblock In {\em Proceedings of the International Conference on Machine
  Learning (ICML)}.

\bibitem[\protect\astroncite{Gulrajani et~al.}{2017}]{gulrajani2017wgangp}
Gulrajani, I., F.~Ahmed, M.~Arjovsky, V.~Dumoulin, and A.~C.
  Courville\leavevmode\nopagebreak\newline 2017.
\newblock Improved training of {Wasserstein} {GANs}.
\newblock In {\em Advances in Neural Information Processing Systems 30}.

\bibitem[\protect\astroncite{Ha and Eck}{2017}]{ha2017neural}
Ha, D. and D.~Eck\leavevmode\nopagebreak\newline 2017.
\newblock A neural representation of sketch drawings.
\newblock \href{https://arxiv.org/abs/1704.03477}{{\ttfamily arXiv:1704.03477
  [cs.NE]}}.

\bibitem[\protect\astroncite{Hertzmann}{2003}]{hertzmann2003survey}
Hertzmann, A.\leavevmode\nopagebreak\newline 2003.
\newblock A survey of stroke-based rendering.
\newblock {\em IEEE Computer Graphics and Applications}, 23:70--81.

\bibitem[\protect\astroncite{Hertzmann}{2018}]{hertzmann2018can}
Hertzmann, A.\leavevmode\nopagebreak\newline 2018.
\newblock Can computers create art?
\newblock {\em Arts}, 7(2):18.

\bibitem[\protect\astroncite{Hoffmann et~al.}{2018}]{hoffmann2018u}
Hoffmann, D.~L., C.~Standish, M.~Garc{\'\i}a-Diez, P.~B. Pettitt, J.~Milton,
  J.~Zilh{\~a}o, J.~J. Alcolea-Gonz{\'a}lez, P.~Cantalejo-Duarte, H.~Collado,
  R.~De~Balb{\'\i}n, and {others}\leavevmode\nopagebreak\newline 2018.
\newblock {U-Th} dating of carbonate crusts reveals {Neandertal} origin of
  {Iberian} cave art.
\newblock {\em Science}, 359(6378):912--915.

\bibitem[\protect\astroncite{Huang et~al.}{2019}]{huang2019learningtopaint}
Huang, Z., W.~Heng, and S.~Zhou\leavevmode\nopagebreak\newline 2019.
\newblock Learning to paint with model-based deep reinforcement learning.
\newblock \href{https://arxiv.org/abs/1903.04411}{{\ttfamily arXiv:1903.04411
  [cs.CV]}}.

\bibitem[\protect\astroncite{Jaderberg et~al.}{2017}]{jaderberg2017population}
Jaderberg, M., V.~Dalibard, S.~Osindero, W.~M. Czarnecki, J.~Donahue,
  A.~Razavi, O.~Vinyals, T.~Green, I.~Dunning, K.~Simonyan, and
  {others}\leavevmode\nopagebreak\newline 2017.
\newblock Population based training of neural networks.
\newblock \href{https://arxiv.org/abs/1711.09846}{{\ttfamily arXiv:1711.09846
  [cs.LG]}}.

\bibitem[\protect\astroncite{Kar et~al.}{2019}]{kar2019meta}
Kar, A., A.~Prakash, M.-Y. Liu, E.~Cameracci, J.~Yuan, M.~Rusiniak, D.~Acuna,
  A.~Torralba, and S.~Fidler\leavevmode\nopagebreak\newline 2019.
\newblock Meta-sim: Learning to generate synthetic datasets.
\newblock \href{https://arxiv.org/abs/1904.11621}{{\ttfamily arXiv:1904.11621
  [cs.CV]}}.

\bibitem[\protect\astroncite{Karras et~al.}{2017}]{karras2017progressive}
Karras, T., T.~Aila, S.~Laine, and J.~Lehtinen\leavevmode\nopagebreak\newline
  2017.
\newblock Progressive growing of gans for improved quality, stability, and
  variation.
\newblock \href{https://arxiv.org/abs/1710.10196}{{\ttfamily arXiv:1710.10196
  [cs.NE]}}.

\bibitem[\protect\astroncite{Karras et~al.}{2018}]{karras2018style}
Karras, T., S.~Laine, and T.~Aila\leavevmode\nopagebreak\newline 2018.
\newblock A style-based generator architecture for generative adversarial
  networks.
\newblock \href{https://arxiv.org/abs/1812.04948}{{\ttfamily arXiv:1812.04948
  [cs.NE]}}.

\bibitem[\protect\astroncite{Kingma and Ba}{2014}]{kingma2014adam}
Kingma, D.~P. and J.~Ba\leavevmode\nopagebreak\newline 2014.
\newblock Adam: A method for stochastic optimization.
\newblock \href{https://arxiv.org/abs/1412.6980}{{\ttfamily arXiv:1412.6980
  [cs.LG]}}.

\bibitem[\protect\astroncite{Kingma and Welling}{2013}]{kingma2013auto}
Kingma, D.~P. and M.~Welling\leavevmode\nopagebreak\newline 2013.
\newblock Auto-encoding variational bayes.
\newblock \href{https://arxiv.org/abs/1312.6114}{{\ttfamily arXiv:1312.6114
  [stat.ML]}}.

\bibitem[\protect\astroncite{Kotani and Tellex}{2019}]{kotani19teaching}
Kotani, A. and S.~Tellex\leavevmode\nopagebreak\newline 2019.
\newblock {Teaching Robots to Draw}.
\newblock In {\em {IEEE International Conference on Robotics and Automation
  (ICRA)}}.

\bibitem[\protect\astroncite{Lake et~al.}{2015}]{lake2015human}
Lake, B.~M., R.~Salakhutdinov, and J.~B.
  Tenenbaum\leavevmode\nopagebreak\newline 2015.
\newblock Human-level concept learning through probabilistic program induction.
\newblock {\em Science}, 350(6266):1332--1338.

\bibitem[\protect\astroncite{Lake et~al.}{2017}]{lake2017building}
Lake, B.~M., T.~D. Ullman, J.~B. Tenenbaum, and S.~J.
  Gershman\leavevmode\nopagebreak\newline 2017.
\newblock Building machines that learn and think like people.
\newblock {\em Behavioral and Brain Sciences}, 40.

\bibitem[\protect\astroncite{L{\'a}zaro-Gredilla
  et~al.}{2019}]{lazaro2018beyond}
L{\'a}zaro-Gredilla, M., D.~Lin, J.~S. Guntupalli, and
  D.~George\leavevmode\nopagebreak\newline 2019.
\newblock Beyond imitation: Zero-shot task transfer on robots by learning
  concepts as cognitive programs.
\newblock {\em Science Robotics}, 4(26):eaav3150.

\bibitem[\protect\astroncite{Li}{2017}]{li2017fluid}
Li, D.\leavevmode\nopagebreak\newline 2017.
\newblock Fluid paint.
\newblock \url{http://david.li/paint/}.

\bibitem[\protect\astroncite{Li et~al.}{2019}]{li2019photosketching}
Li, M., Z.~Lin, R.~Mech, E.~Yumer, and
  D.~Ramanan\leavevmode\nopagebreak\newline 2019.
\newblock Photo-sketching: Inferring contour drawings from images.
\newblock In {\em WACV 2019: IEEE Winter Conf. on Applications of Computer
  Vision}.

\bibitem[\protect\astroncite{Liu et~al.}{2019}]{liu2018learning}
Liu, Y., Z.~D. Wu, D.~A. Ritchie, W.~T. Freeman, J.~B. Tenenbaum, and
  J.~Wu\leavevmode\nopagebreak\newline 2019.
\newblock Learning to describe scenes with programs.
\newblock In {\em ICLR}.

\bibitem[\protect\astroncite{Louppe et~al.}{2019}]{louppe2017adversarial}
Louppe, G., J.~Hermans, and K.~Cranmer\leavevmode\nopagebreak\newline 2019.
\newblock Adversarial variational optimization of non-differentiable
  simulators.
\newblock In {\em The 22nd International Conference on Artificial Intelligence
  and Statistics}, Pp.~ 1438--1447.

\bibitem[\protect\astroncite{Minsky and Papert}{1972}]{minsky1972artificial}
Minsky, M. and S.~A. Papert\leavevmode\nopagebreak\newline 1972.
\newblock Artificial intelligence progress report.

\bibitem[\protect\astroncite{Miyato et~al.}{2018}]{miyato2018spectral}
Miyato, T., T.~Kataoka, M.~Koyama, and
  Y.~Yoshida\leavevmode\nopagebreak\newline 2018.
\newblock Spectral normalization for generative adversarial networks.
\newblock In {\em International Conference on Learning Representations}.

\bibitem[\protect\astroncite{Nakano}{2019}]{nakano2019neural}
Nakano, R.\leavevmode\nopagebreak\newline 2019.
\newblock Neural painters: A learned differentiable constraint for generating
  brushstroke paintings.
\newblock \href{https://arxiv.org/abs/1904.08410}{{\ttfamily arXiv:1904.08410
  [cs.CV]}}.

\bibitem[\protect\astroncite{Ng et~al.}{1999}]{ng1999rewardtransformations}
Ng, A.~Y., D.~Harada, and S.~J. Russell\leavevmode\nopagebreak\newline 1999.
\newblock Policy invariance under reward transformations: Theory and
  application to reward shaping.
\newblock In {\em Proceedings of the Sixteenth International Conference on
  Machine Learning (ICML 1999)}.

\bibitem[\protect\astroncite{Pathak et~al.}{2016}]{pathak2016inpainting}
Pathak, D., P.~Krahenbuhl, J.~Donahue, T.~Darrell, and A.~A.
  Efros\leavevmode\nopagebreak\newline 2016.
\newblock Context encoders: Feature learning by inpainting.
\newblock {\em 2016 IEEE Conference on Computer Vision and Pattern Recognition
  (CVPR)}.

\bibitem[\protect\astroncite{Renold}{2004}]{renold2004mypaint}
Renold, M.\leavevmode\nopagebreak\newline 2004.
\newblock Mypaint.
\newblock \url{http://mypaint.org/}.

\bibitem[\protect\astroncite{Ruiz et~al.}{2018}]{ruiz2018learning}
Ruiz, N., S.~Schulter, and M.~Chandraker\leavevmode\nopagebreak\newline 2018.
\newblock Learning to simulate.
\newblock \href{https://arxiv.org/abs/1810.02513}{{\ttfamily arXiv:1810.02513
  [cs.LG]}}.

\bibitem[\protect\astroncite{Russakovsky
  et~al.}{2015}]{russakovsky2015imagenet}
Russakovsky, O., J.~Deng, H.~Su, J.~Krause, S.~Satheesh, S.~Ma, Z.~Huang,
  A.~Karpathy, A.~Khosla, M.~Bernstein, A.~C. Berg, and
  L.~Fei-Fei\leavevmode\nopagebreak\newline 2015.
\newblock {ImageNet Large Scale Visual Recognition Challenge}.
\newblock {\em International Journal of Computer Vision (IJCV)},
  115(3):211--252.

\bibitem[\protect\astroncite{Schmidhuber}{1997}]{schmidhuber1997low}
Schmidhuber, J.\leavevmode\nopagebreak\newline 1997.
\newblock Low-complexity art.
\newblock {\em Leonardo}, 30(2):97--103.

\bibitem[\protect\astroncite{Selim}{2018}]{selim2018spiderman}
Selim, M.\leavevmode\nopagebreak\newline 2018.
\newblock Spiderman 10 min 1 min 10 sec speed challenge.
\newblock \url{https://www.youtube.com/watch?v=x9wn633vl_c}.

\bibitem[\protect\astroncite{Sutton}{2019}]{sutton2019bitter}
Sutton, R.\leavevmode\nopagebreak\newline 2019.
\newblock The bitter lesson.
\newblock \url{http://www.incompleteideas.net/IncIdeas/BitterLesson.html}.

\bibitem[\protect\astroncite{van~den Oord et~al.}{2016}]{oord2016pixel}
van~den Oord, A., N.~Kalchbrenner, and
  K.~Kavukcuoglu\leavevmode\nopagebreak\newline 2016.
\newblock Pixel recurrent neural networks.
\newblock In {\em International Conference on Machine Learning}, Pp.~
  1747--1756.

\bibitem[\protect\astroncite{Welling}{2019}]{welling2019do}
Welling, M.\leavevmode\nopagebreak\newline 2019.
\newblock Do we still need models or just more data and compute?
\newblock
  \url{https://staff.fnwi.uva.nl/m.welling/wp-content/uploads/Model-versus-Data-AI-1.pdf}.

\bibitem[\protect\astroncite{Zeng et~al.}{2009}]{zeng2009image}
Zeng, K., M.~Zhao, C.~Xiong, and S.-C. Zhu\leavevmode\nopagebreak\newline 2009.
\newblock From image parsing to painterly rendering.
\newblock {\em ACM Trans. Graph.}, 29:2:1--2:11.

\bibitem[\protect\astroncite{Zheng et~al.}{2019}]{zheng2018strokenet}
Zheng, N., Y.~Jiang, and D.~jiang Huang\leavevmode\nopagebreak\newline 2019.
\newblock Strokenet: A neural painting environment.
\newblock In {\em ICLR}.

\bibitem[\protect\astroncite{Zhou et~al.}{2018}]{zhou2018learning}
Zhou, T., C.~Fang, Z.~Wang, J.~Yang, B.~Kim, Z.~Chen, J.~Brandt, and
  D.~Terzopoulos\leavevmode\nopagebreak\newline 2018.
\newblock Learning to sketch with deep {Q} networks and demonstrated strokes.
\newblock \href{https://arxiv.org/abs/1810.05977}{{\ttfamily arXiv:1810.05977
  [cs.CV]}}.

\end{thebibliography}

\clearpage

\begin{appendices}

\section{Compound action space}
\label{sect:compound_action_space}

In \cite{ganin2018spiral}, a new stroke is deposited onto the canvas at every step of the environment. Consequently, each stroke has a fixed quadratic B\'ezier form with start coordinates, mid-point coordinates and end coordinates (in addition to thickness and colour). The agent produces new values for the mid-point and end coordinates at every step and selects a thickness and colour for the stroke. The previous step's end coordinate is used for the current step's start coordinate. We refer to this interface as the \textit{simple} interface.

In this work, we also investigate a \textit{compound} interface, where each stroke is built up over the course of several environment steps. In each step, the agent produces the coordinates of a new control point in the current stroke. In order to terminate a stroke, the agent can execute a discrete action, at which point the stroke is deposited onto the canvas as a cubic spline going through the control points. The thickness and colour of the stroke are determined at the beginning of a sequence of strokes. The compound interface allows the agents to produce smoother, more precise curves, which can sometimes lead to more aesthetically pleasing results. The compound interface is also instrumental in parsing Omniglot characters, as we show in \autoref{sect:results_omniglot}.

\FloatBarrier\section{Sample diversity}

\begin{figure}[ht]
\centering
\begin{subfigure}[b]{0.32\textwidth}
    \adjincludegraphics[width=\textwidth, trim={0 {.25\height} {.5\width} 0}, clip]{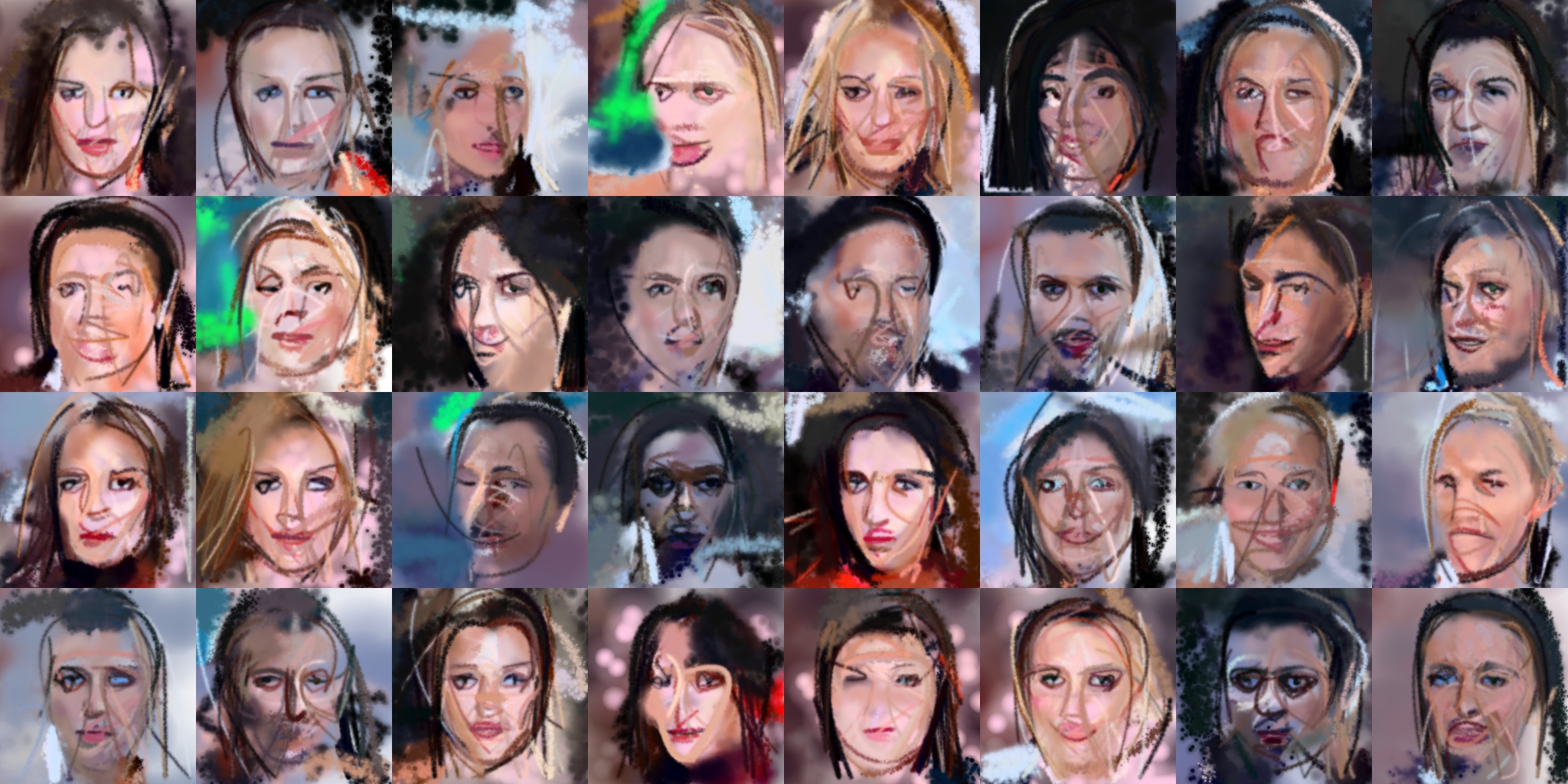}
\end{subfigure}
\hfill
\begin{subfigure}[b]{0.32\textwidth}
    \adjincludegraphics[width=\textwidth, trim={0 {.25\height} {.5\width} 0}, clip]{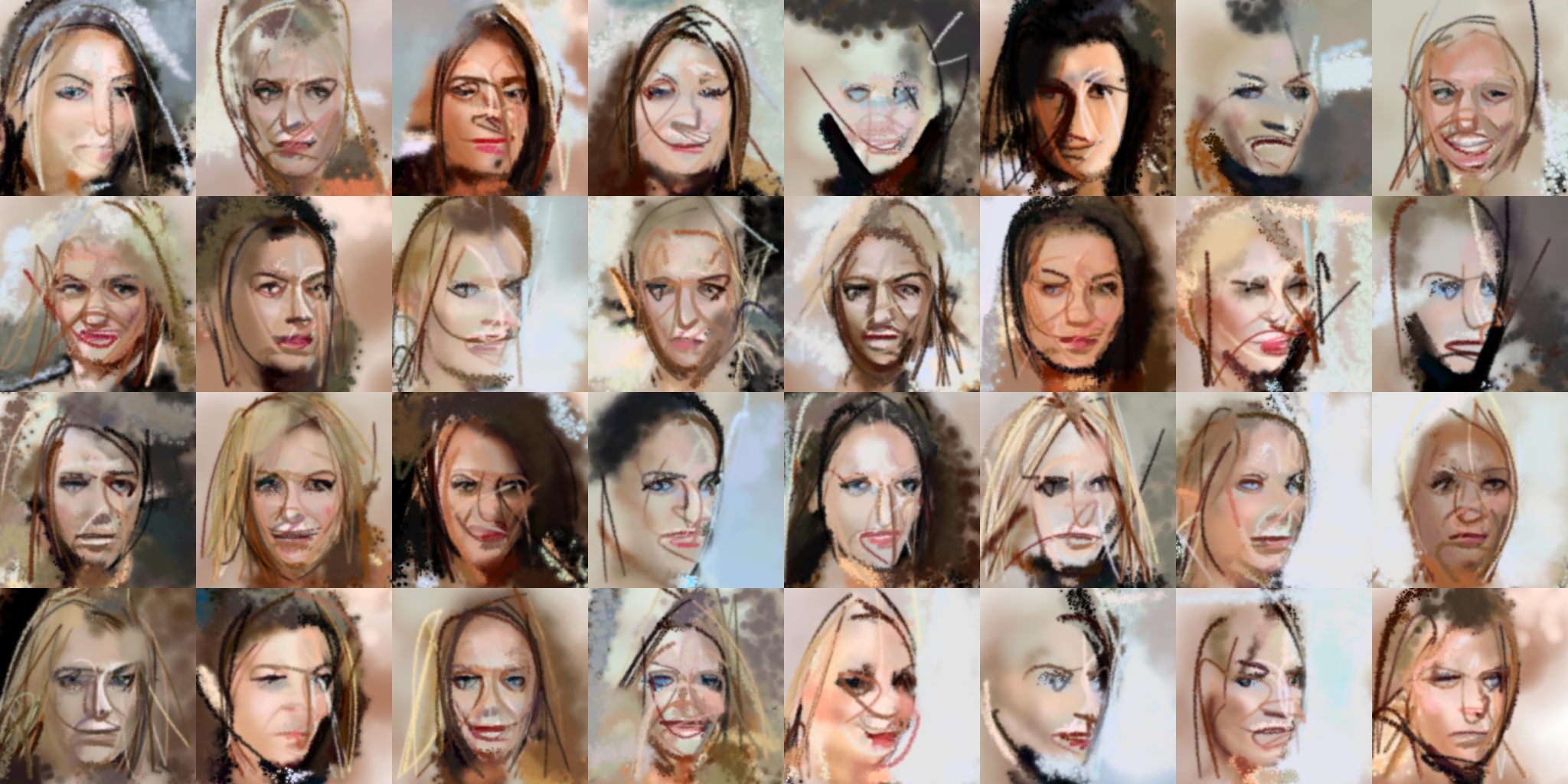}
\end{subfigure}
\hfill
\begin{subfigure}[b]{0.32\textwidth}
    \adjincludegraphics[width=\textwidth, trim={0 {.25\height} {.5\width} 0}, clip]{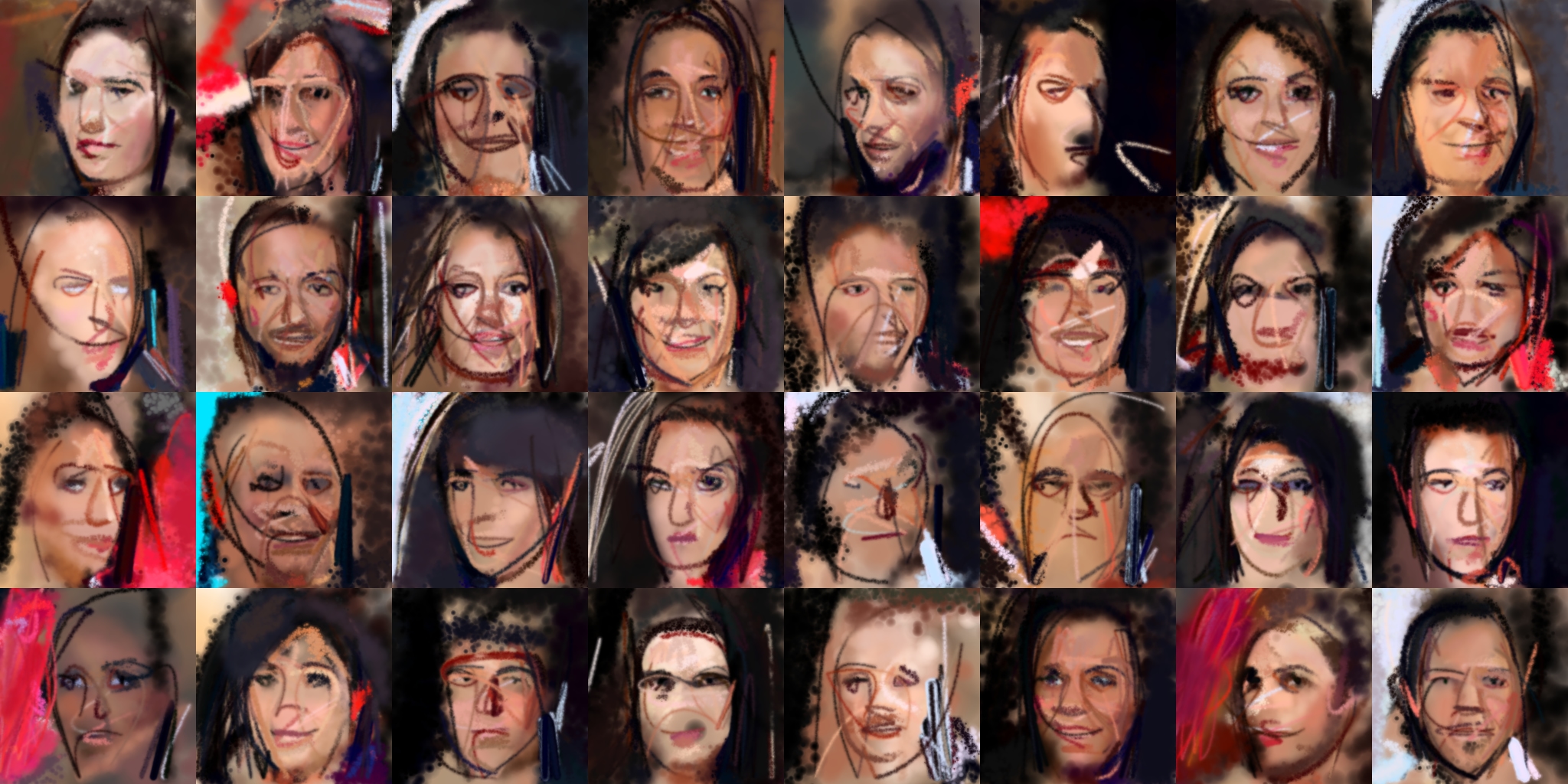}
\end{subfigure}
\caption{\textbf{Within-population and within-agent diversity (long episodes).} We show three sets of samples for three different agents in the same population, showing both the diversity of samples for each agent, and the diversity of samples across agents in the same population. Mode collapse occurs less frequently in long episodes. \label{fig:gen_200steps_multi}}
\end{figure}

\begin{figure}[h]
\centering
\includegraphics[height=0.16\textwidth]{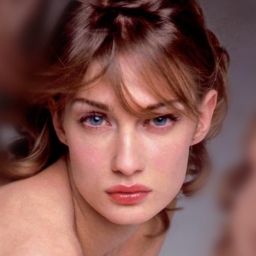}
\hfill
\includegraphics[height=0.16\textwidth]{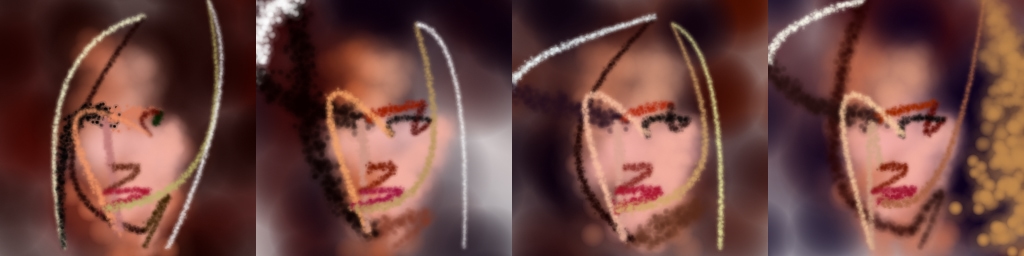}
\hfill
\includegraphics[height=0.16\textwidth]{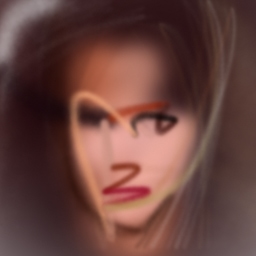}
\caption{\textbf{Conditional generation stochasticity.} We visualise an agent's stochasticity by producing multiple samples for the same target image. Agent stochasticity is due to each action being sampled auto-regressively. Left: Target image. Middle: Multiple samples from the same agent. Right: Mean image produced by averaging 100 samples.}
\label{fig:rec_fixed_target}
\end{figure}

\autoref{fig:gen_20steps_multi} and \autoref{fig:gen_200steps_multi} show diversity of unconditional samples. \autoref{fig:rec_fixed_target} shows that there is also considerable diversity in reconstructions of the same target.


\clearpage\section{Evaluation on ImageNet}
\label{sect:results_imagenet}

We also try the \generativeagentsname framework on the ImageNet dataset \citep{russakovsky2015imagenet} to see if it generalizes to more varied images that have not been aligned/cropped. However to keep the task simpler we train on individual classes from ImageNet, so only about 1000 images are used for training each agent, again downscaled to 64 $\times$ 64.

We sometimes instead train using an infinite dataset of samples for single ImageNet classes generated by a pre-trained BigGAN \citep{brock2018biggan} with a truncation threshold of 0.4. This is similar to training on ImageNet directly, but due to the truncation the samples generated are less diverse: objects tend to be centered, face on, with normal appearances and less cluttered backgrounds.

\begin{figure}[h]
\centering
\includegraphics[width=0.32\textwidth]{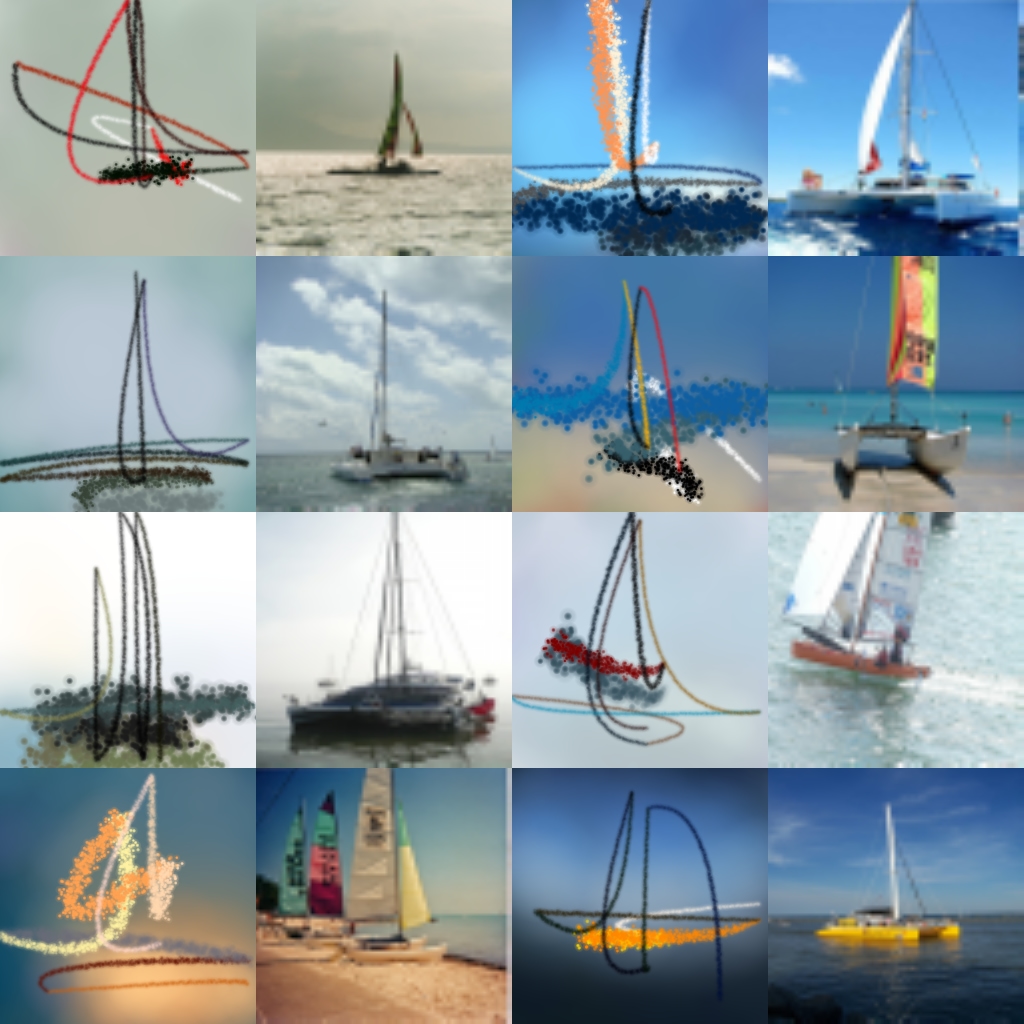}
\hfill
\includegraphics[width=0.32\textwidth]{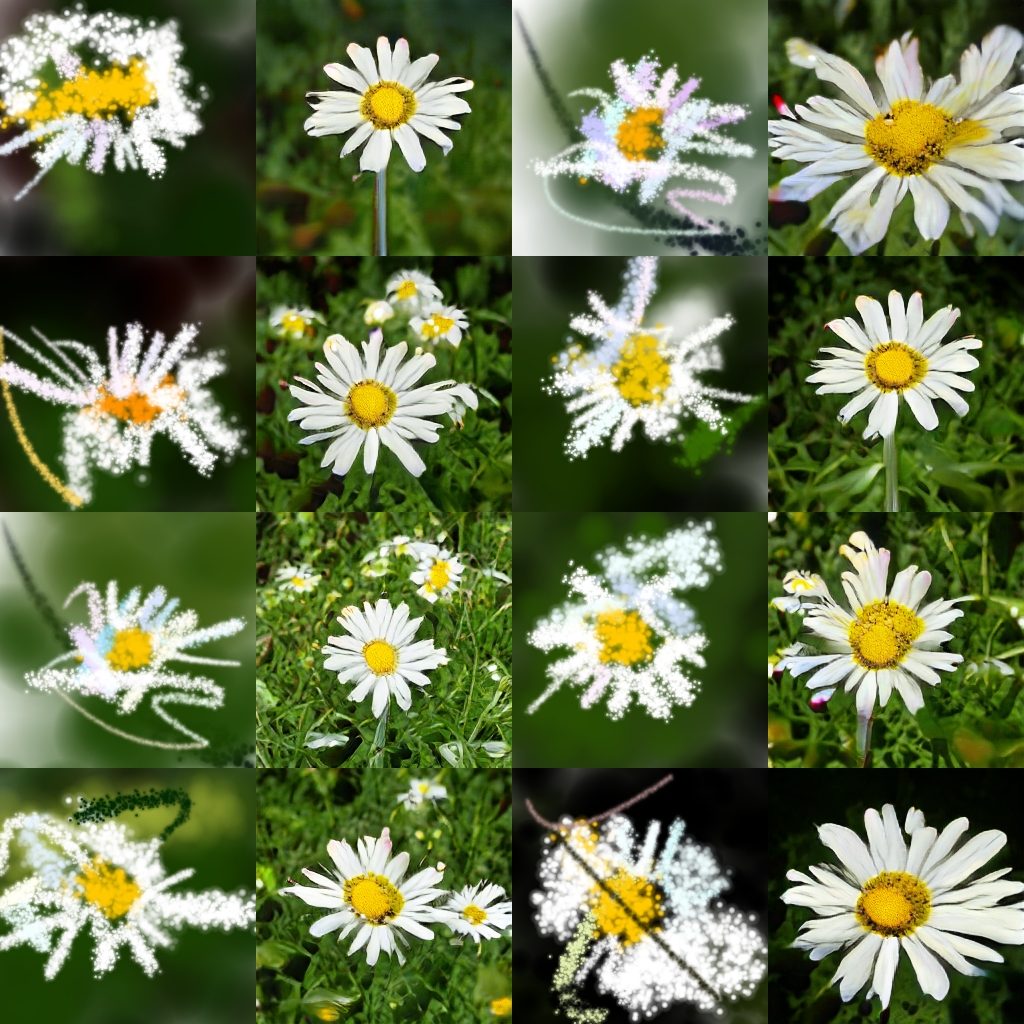}
\hfill
\includegraphics[width=0.32\textwidth]{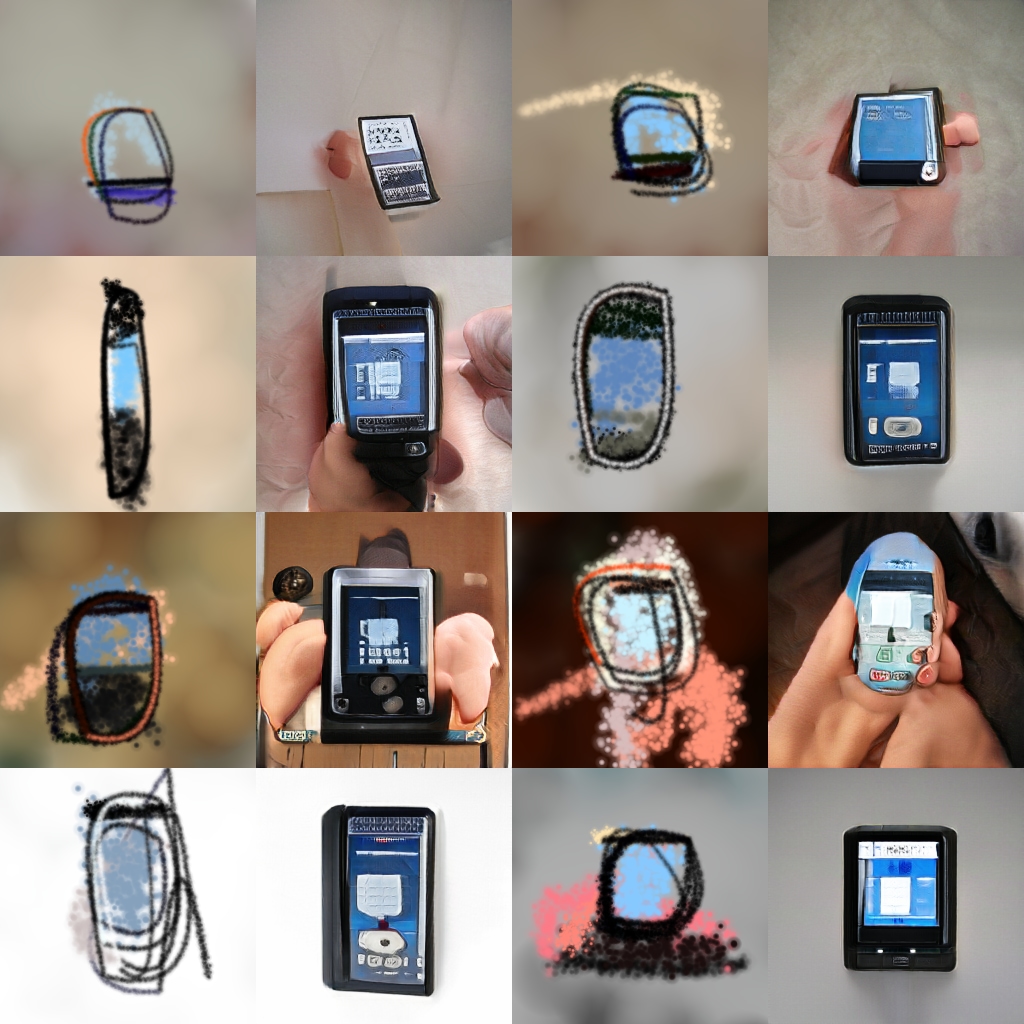}
\caption{\textbf{Cherry-picked samples of reconstruction with complement discriminator, trained for 20 steps on single classes of ImageNet.} Left: Trained on ImageNet catamaran class. Middle: Trained on BigGAN samples of daisies. Right: Trained on BigGAN samples of phones.}
\label{fig:imagenet_classes}
\end{figure}

In both cases \generativeagentsname are able to reproduce the rough colors, and sometimes as in \autoref{fig:imagenet_classes} can extract high-level structure (catamarans have masts, daisies have radial lines, and so on), but results are less reliable than for faces.

\clearpage\section{Evaluation on Omniglot}
\label{sect:results_omniglot}

\begin{figure}[h]
\centering
\includegraphics[height=0.22\textheight]{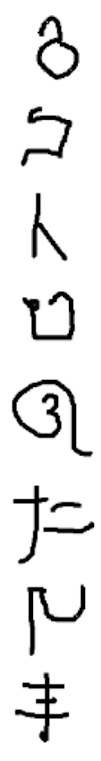}
\includegraphics[height=0.22\textheight]{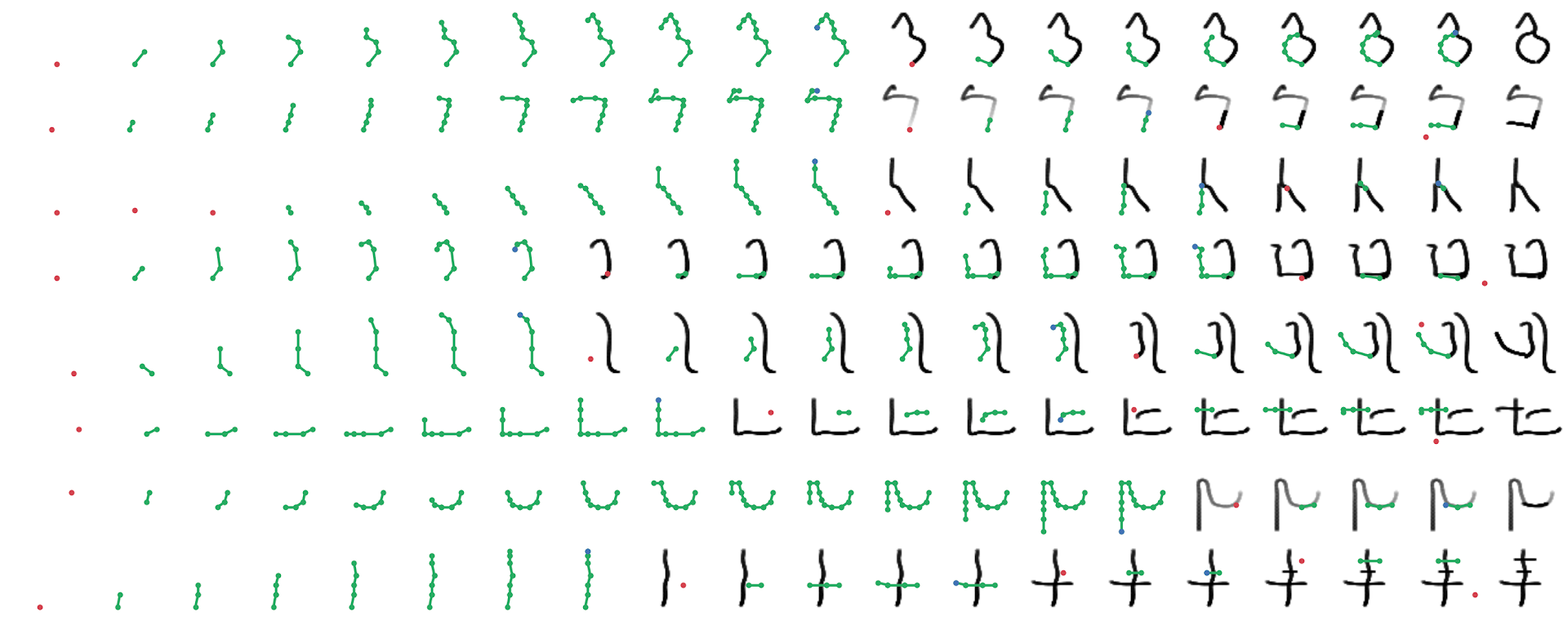}
\caption{\textbf{Omniglot parsing.} Left: Target images. Right: Sequence of actions taken by a single agent to recreate the target images. Red dots signify a request from the agent to commit the current compound stroke to the canvas and to start a new stroke. Green dots indicate locations of commands that build up a compound stroke. The agent is mostly successful in parsing characters into plausible strokes. At times, it fails to complete the character in the allocated episode length.}
\label{fig:omniglot_rec_filmstripsheet}
\end{figure}

\begin{figure}[h]
\centering
\includegraphics[width=0.45\textwidth]{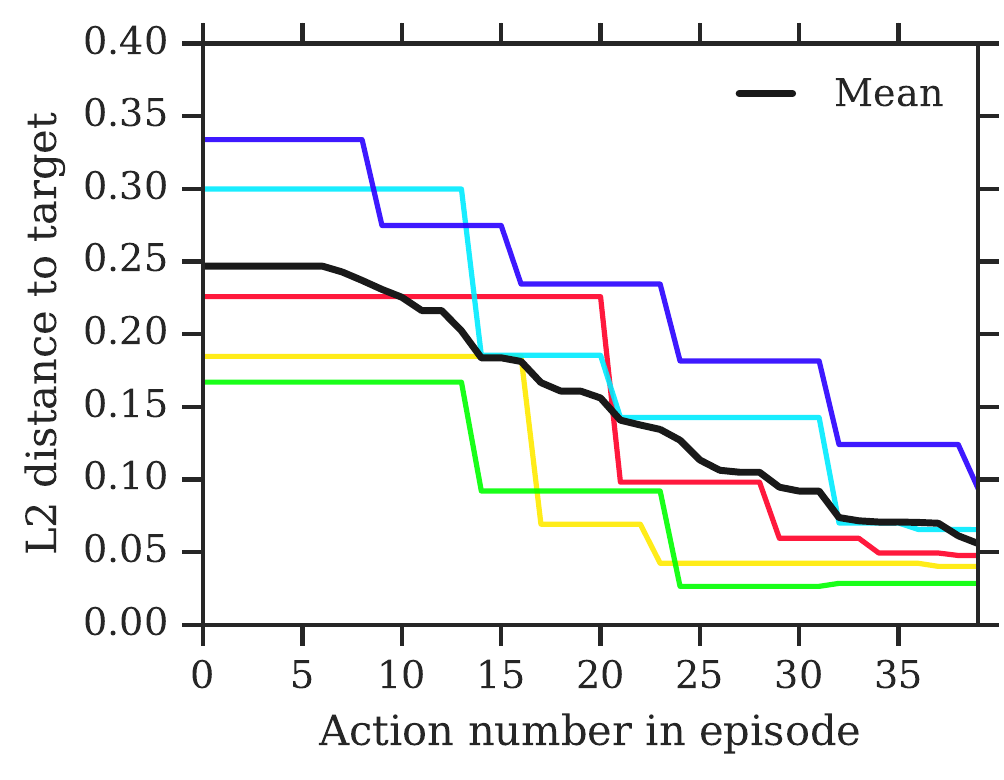}
\hfill
\includegraphics[width=0.45\textwidth]{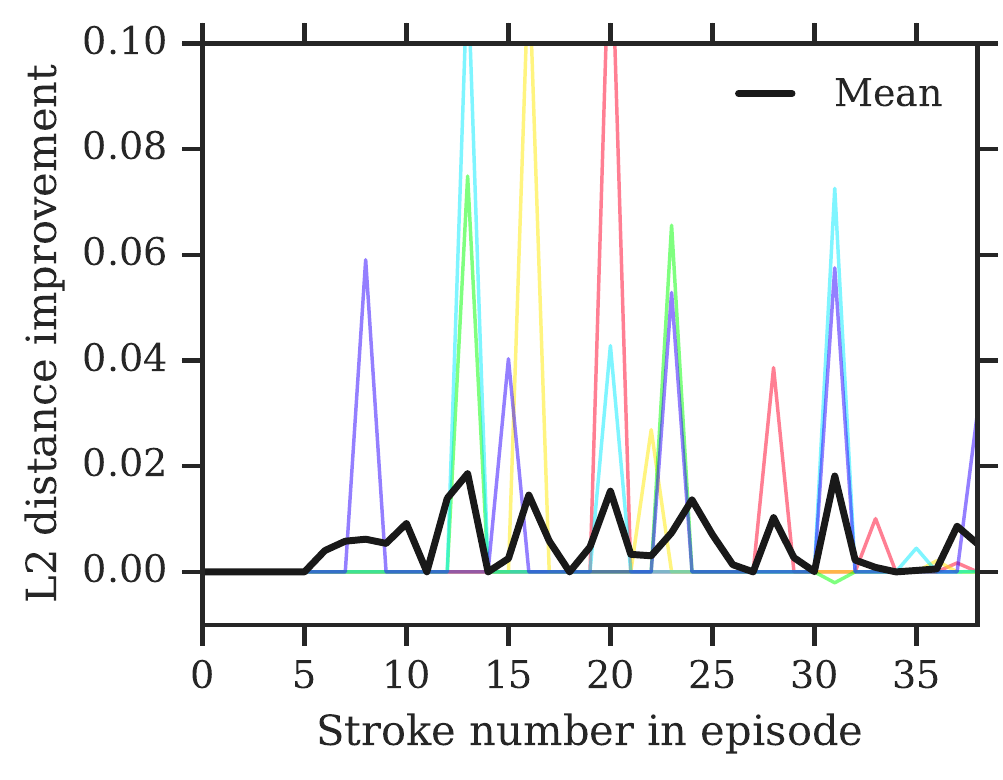}
\caption{\textbf{Within episode `improvement' of image quality (Omniglot).} Left: We plot the L2 distance of the canvas to its target image for 5 different episodes of a 40 step agent (each curve a different episode). Right: The amount of improvement in L2 distance in each step.}
\label{fig:omniglot_rec_distances}
\end{figure}

We also experiment with applying the \generativeagentsname framework to the Omniglot dataset \citep{lake2015human}. In this setting, we train conditional \generativeagentsname to reconstruct Omniglot characters, thereby learning a mapping from bitmap representations into strokes. Note that we do not use the human stroke data for training. We primarily use agents with compound action spaces (\autoref{sect:compound_action_space}) to allow them to express more complex, intricately curved strokes when parsing. We additionally introduce a number of hyperparameters to encourage the agent to learn more natural behaviours. In particular, the environment rewards the agent with a small penalty for each time a new stroke is created, as well as a penalty on the total length of each stroke. Without these penalties, agents achieve almost perfect reconstructions but with un-natural movements. We set the values of these parameters via trial and error and visual inspection. A sample of these results can be seen in \autoref{fig:omniglot_rec_filmstripsheet} and \autoref{fig:omniglot_rec_distances}.

\clearpage\section{Training curves}

\begin{figure}[h]
\centering
\includegraphics[height=0.35\textwidth]{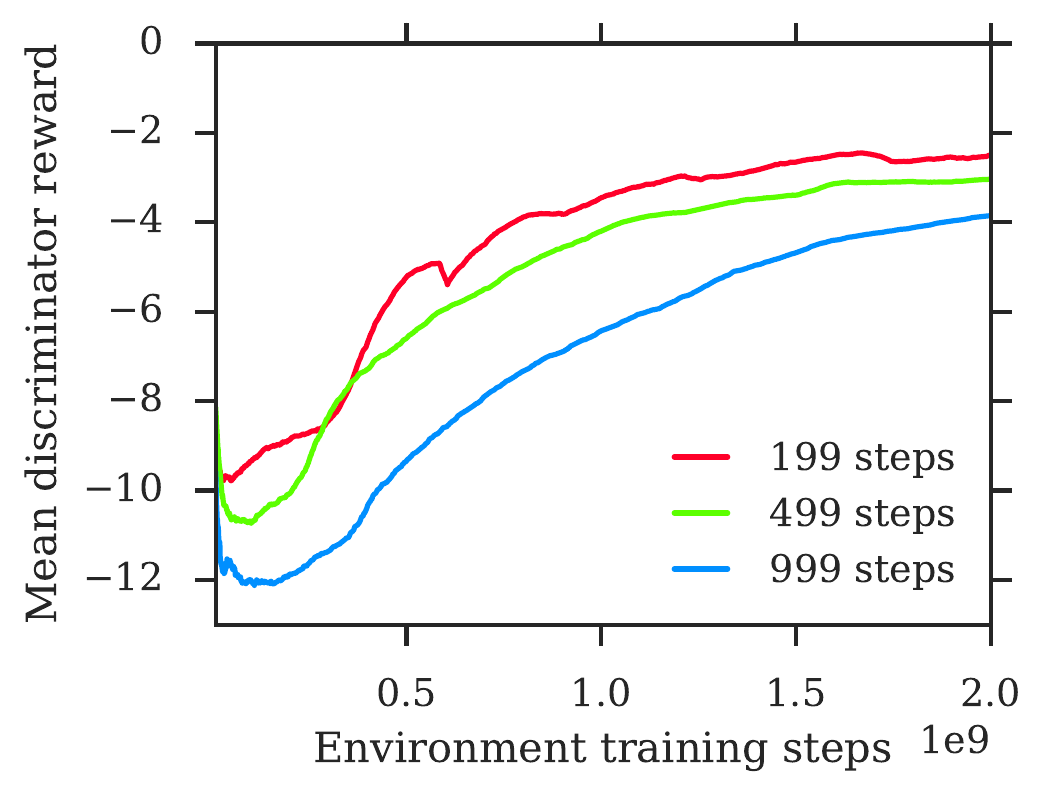}
\hfill
\includegraphics[height=0.35\textwidth]{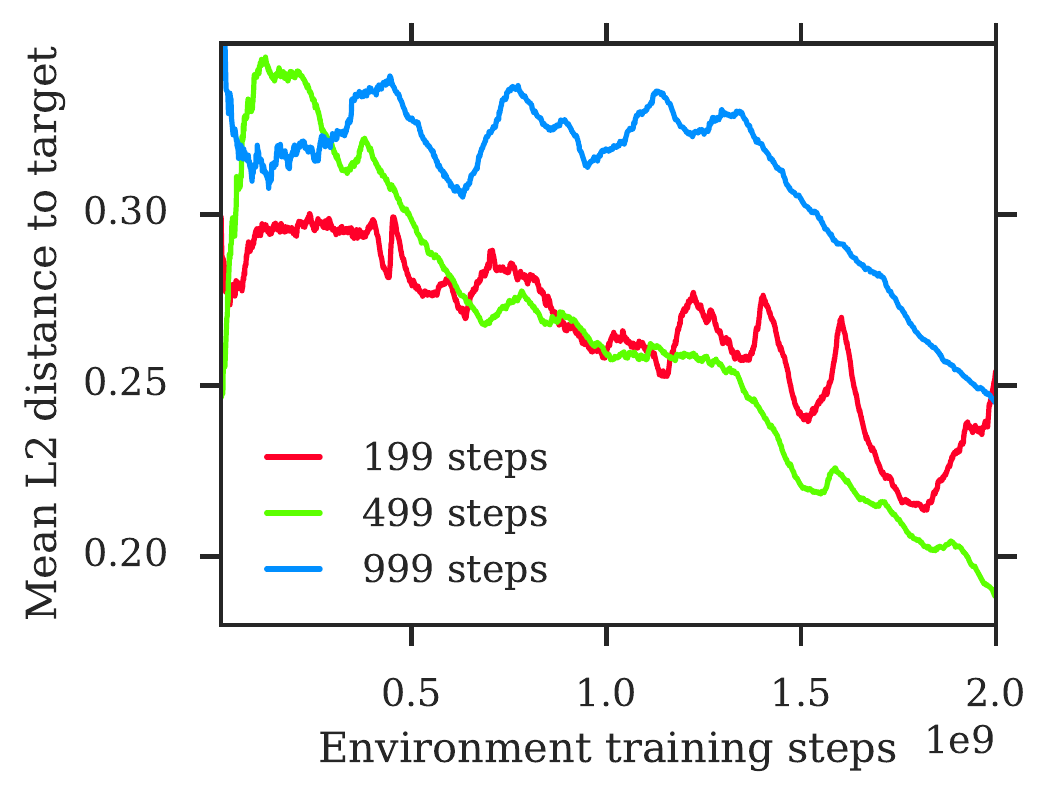}
\caption{\textbf{Conditional training curves with long episodes.} Left: Mean discriminator reward achieved by the generators on the final step of the episode as training progresses. Note that the absolute value of the curves cannot be compared across different agents, due to each competing with their own discriminators. Right: Mean L2 distance of the final generated image and the target image. Note that the agents are trained to maximise the discriminator reward, but we observe some correlation between that objective and minimisation of L2 error.}
\label{fig:rec_training}
\end{figure}

\begin{figure}[h]
\centering
\includegraphics[height=0.35\textwidth]{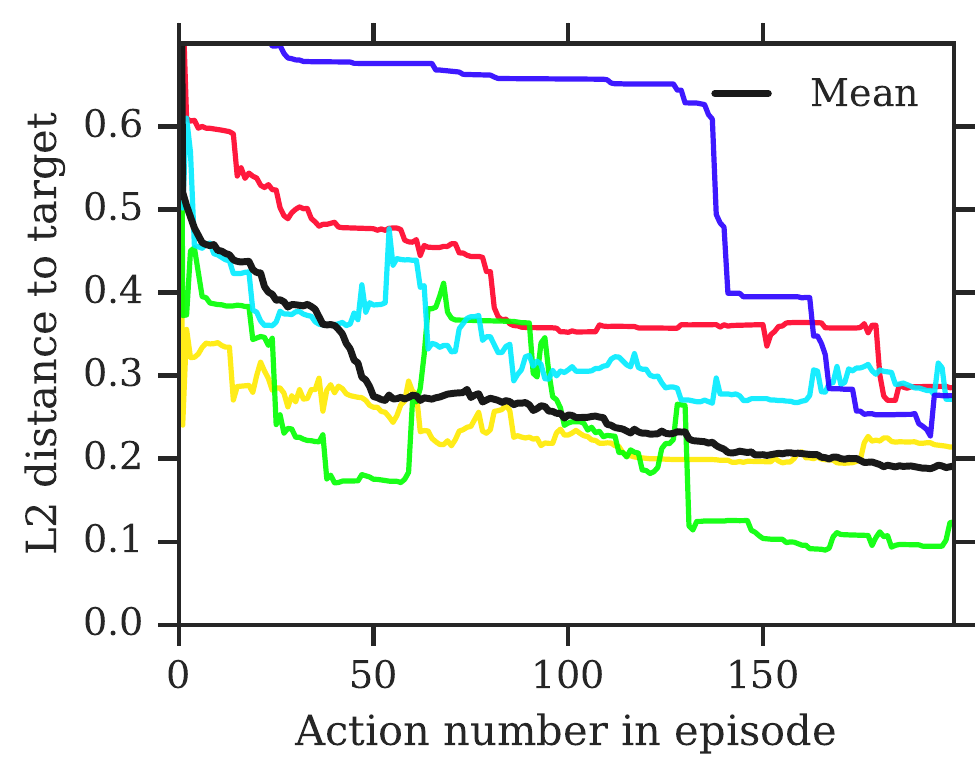}
\hfill
\includegraphics[height=0.35\textwidth]{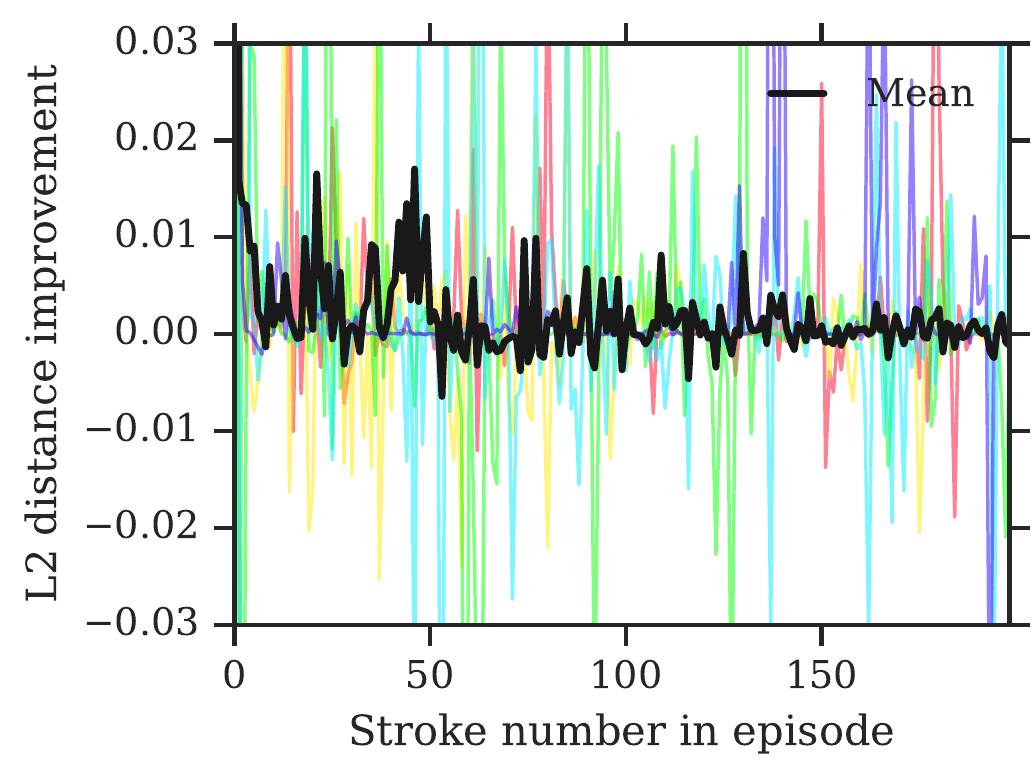}
\caption{\textbf{Within episode `improvement' of image quality (CelebA).} Left: We plot the L2 distance of the canvas to its target image for 5 different episodes of a conditional 200 step agent trained to reconstruct specific target images with discount factor 0.99 (each curve a different episode). Right: The amount of improvement in L2 distance in each step. We see that the agent often takes actions that increase the distance between the canvas and the target.}
\label{fig:rec_distances}
\end{figure}

\autoref{fig:rec_training}, \autoref{fig:rec_distances} show training curves for representative agents on the CelebA-HQ dataset.


\clearpage\section{Ablations}
\label{sect:ablations}

\begin{figure}[h]
\centering
\begin{subfigure}[b]{\textwidth}
    \begin{subfigure}[b]{0.49\textwidth}
        \adjincludegraphics[width=\textwidth, trim={{.5\width} {.25\width} 0 0}, clip]{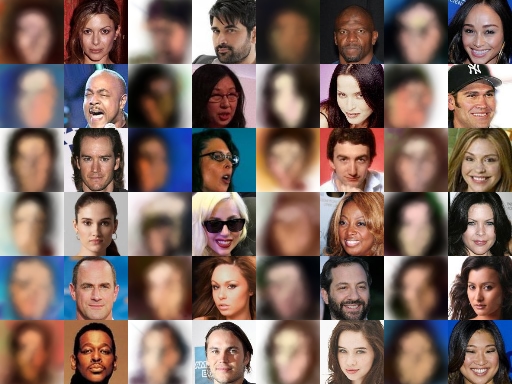}
        \caption{Conditional \oldmodelname}
    \end{subfigure}
    \hfill
    \begin{subfigure}[b]{0.49\textwidth}
        \includegraphics[width=\textwidth]{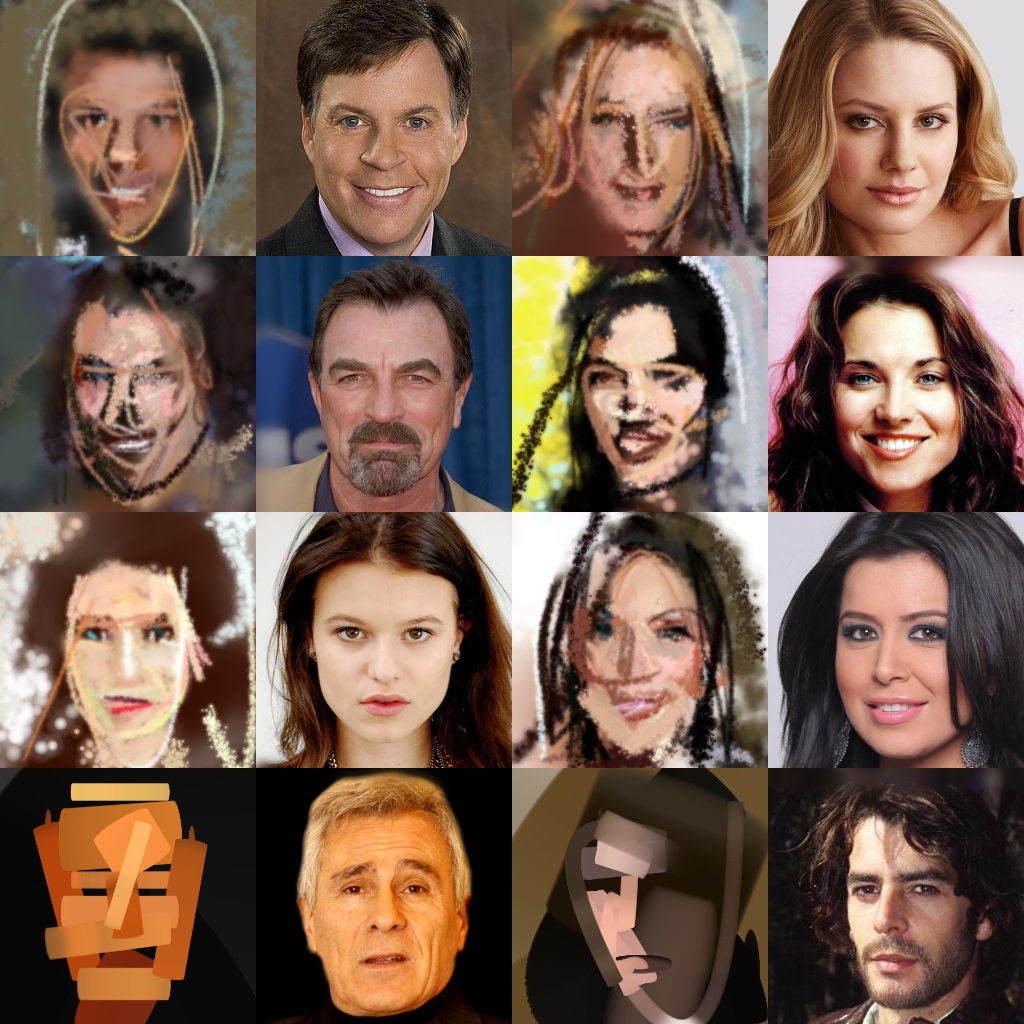}
        \caption{Conditional \newmodelname}
    \end{subfigure}
\end{subfigure}
\vfill
\begin{subfigure}[b]{\textwidth}
    \begin{subfigure}[b]{\textwidth}
        \adjincludegraphics[width=0.49999\textwidth, trim={0 {.501\height} 0 0}, clip]{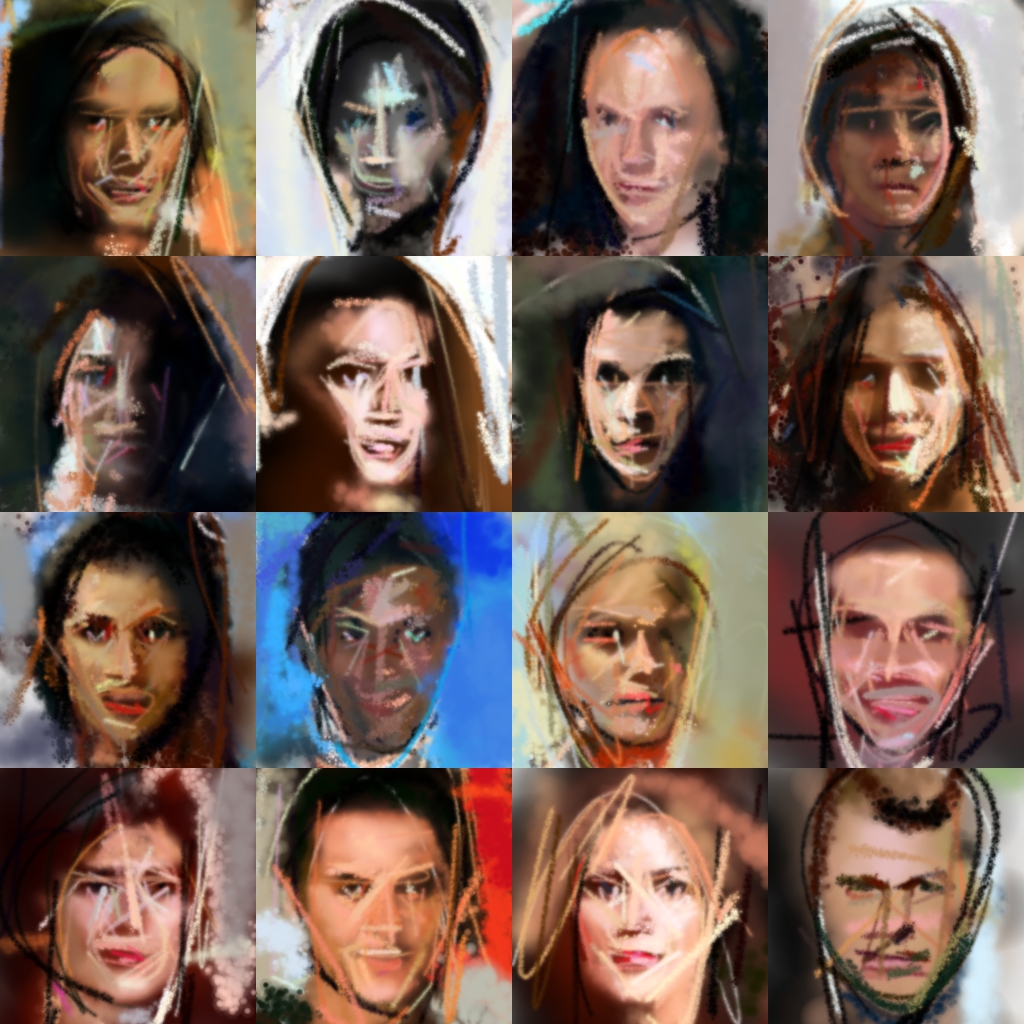}%
        \adjincludegraphics[width=0.49999\textwidth, trim={0 0 0 {.501\height}}, clip]{im/improvements-unconditional2}
        \caption{Unconditional \newmodelname}
    \end{subfigure}
\end{subfigure}
\caption{\textbf{Quality improvements over \oldmodelname.}
(a)~Reconstructions taken from \citep{ganin2018spiral}. Note that these results are for CelebA (not CelebA-HQ used in the present paper). Unfortunately \citep{ganin2018spiral} did not include generation results.
(b)~Selected reconstructions with our improvements.
(c)~Selected unconditional generation with our improvements.}
\label{fig:improvements}
\end{figure}

We provide ablations to show the impact of our design choices. \autoref{fig:improvements} shows the combined improvement from all our modifications detailed under \autoref{sect:spiral_plusplus_improvements}. The following sections separately ablate important modifications.

\FloatBarrier\inlinesubsection{Spectral Normalization ablation}
\label{sect:disc_reg_ablation}

\begin{figure}[htbp]
\centering
\begin{subfigure}[b]{0.49\textwidth}
    \includegraphics[width=\textwidth]{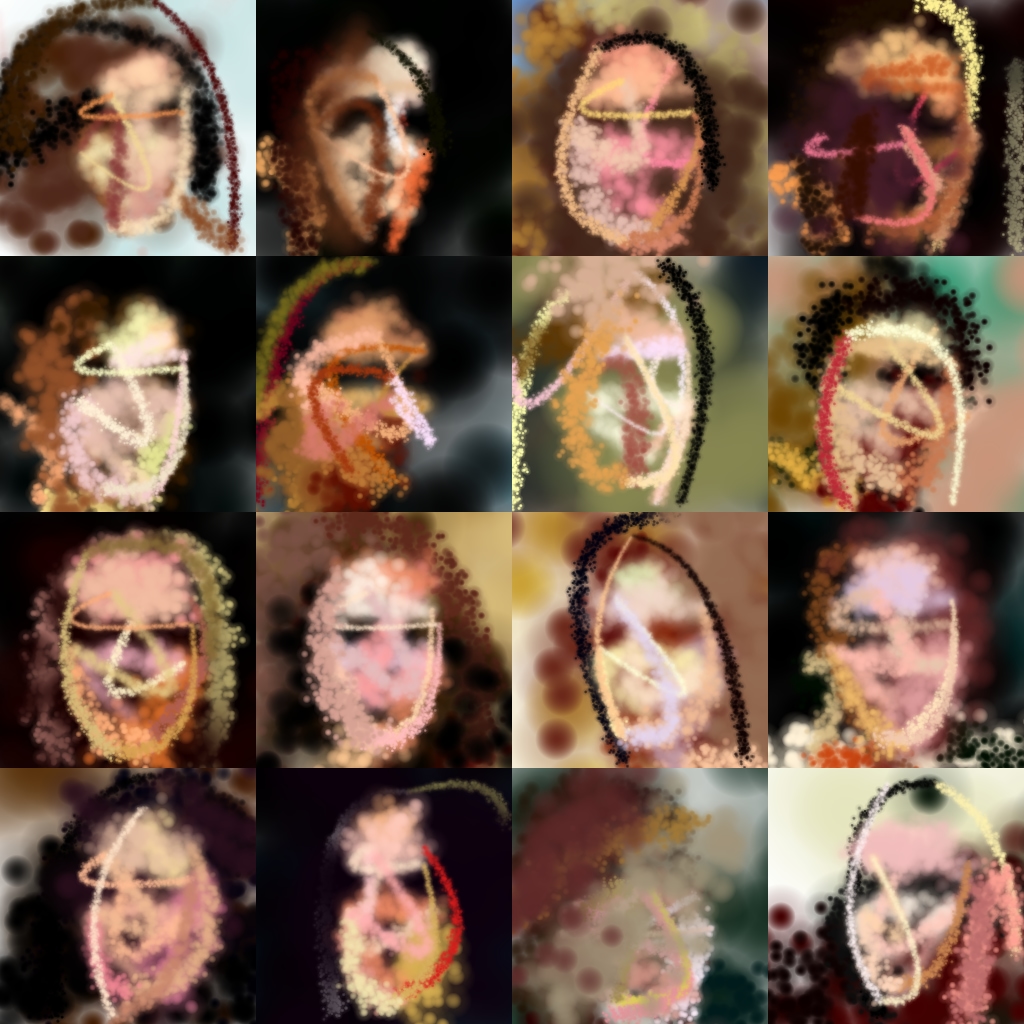}
    \caption{Gradient penalty (WGAN-GP)}
\end{subfigure}
\hfill
\begin{subfigure}[b]{0.49\textwidth}
    \includegraphics[width=\textwidth]{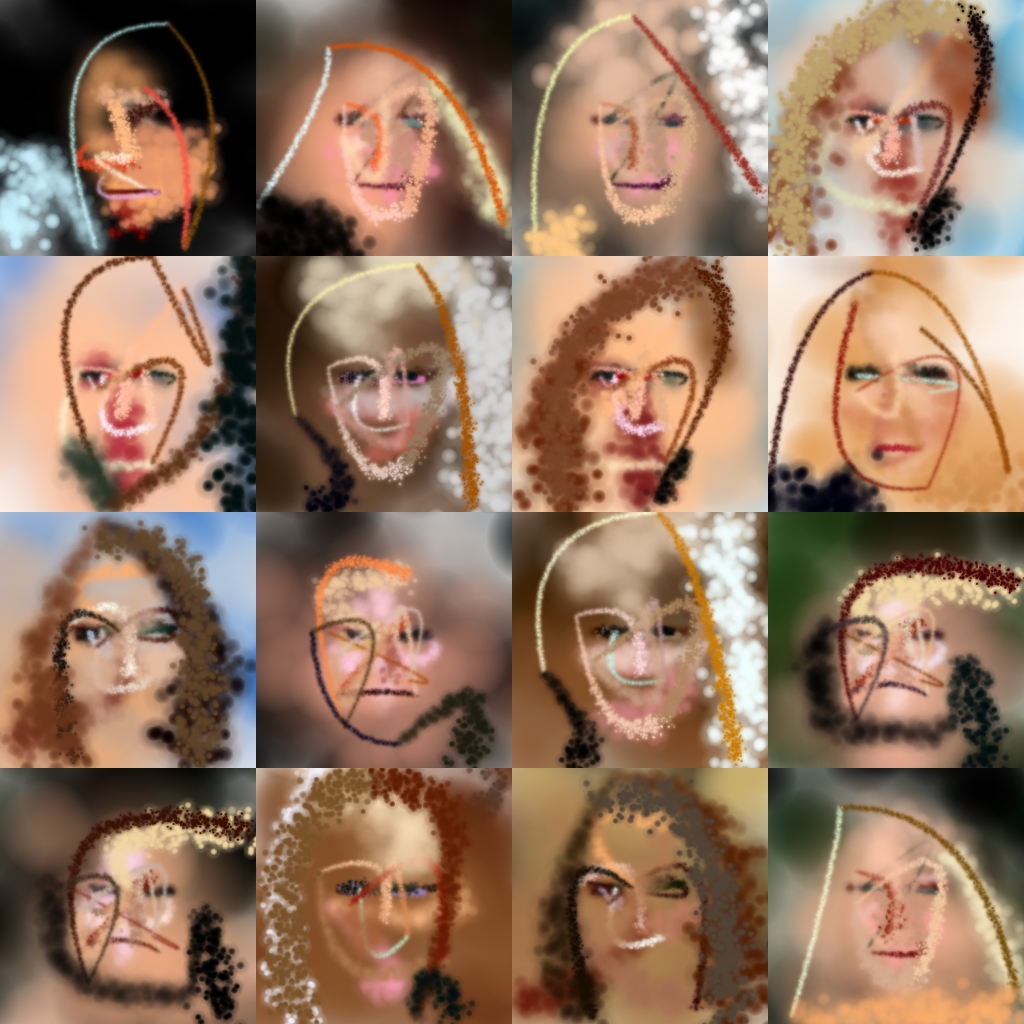}
    \caption{Spectral normalization}
\end{subfigure}
\caption{\textbf{Comparison of discriminator regularization on 19 step episodes.} Spectral Normalization lets the discriminator guide the agent into producing finer details like eyes, noses and mouths. These samples are from a single agent in the population; \autoref{fig:gen_20steps_multi} shows how different agents in the same population focus on reproducing different aspects of the target image distribution.}
\label{fig:sn_ablation}
\end{figure}

\autoref{fig:sn_ablation} shows the impact of the discriminator regularization modifications in \autoref{sect:spectral_norm}.

\inlinesubsection{Temporal Credit Assignment ablation}
\label{sect:ca_ablation}

\begin{figure}[htbp]
\centering
\adjincludegraphics[width=\textwidth, trim={0 {.2501\width} 0 0}, clip]{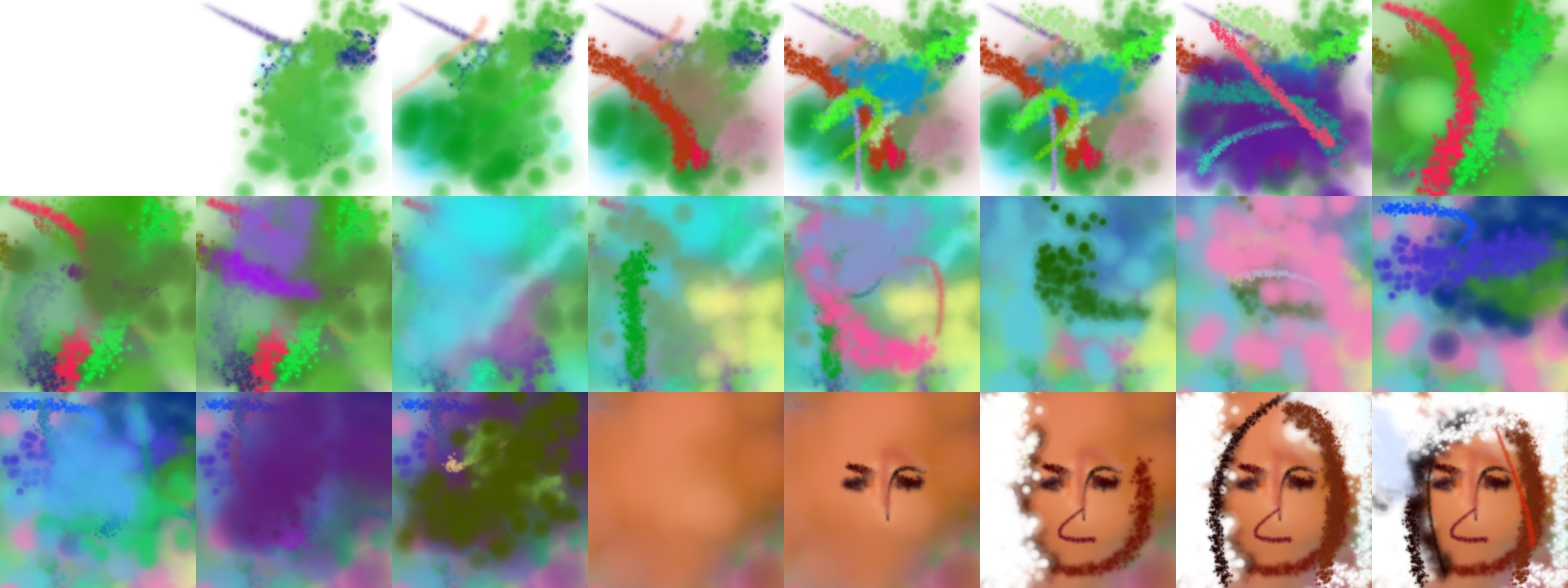}
\vskip 2mm
\adjincludegraphics[width=\textwidth, trim={0 {.1251\width} 0 {.1251\width}}, clip]{im/noca_gen_200steps_filmstrip_every_8_and_a_half_frames_3rows}
\vskip 2mm
\adjincludegraphics[width=\textwidth, trim={0 0 0 {.2501\width}}, clip]{im/noca_gen_200steps_filmstrip_every_8_and_a_half_frames_3rows}
\caption{\textbf{Agents without TCA often waste actions.} These are 24 equally spaced frames taken from a single episode with 199 steps. The agent wastes its first 156 actions drawing brightly colored strokes that it eventually paints over, finally drawing a rough face in the final 43 steps. By comparison, agents with TCA start drawing immediately and gradually refine the image, as in \autoref{fig:gen_1000step_filmstrips}.}
\label{fig:noca_200steps}
\end{figure}

\begin{figure}[htbp]
\centering
\begin{subfigure}[b]{0.49\textwidth}
    \includegraphics[width=\textwidth]{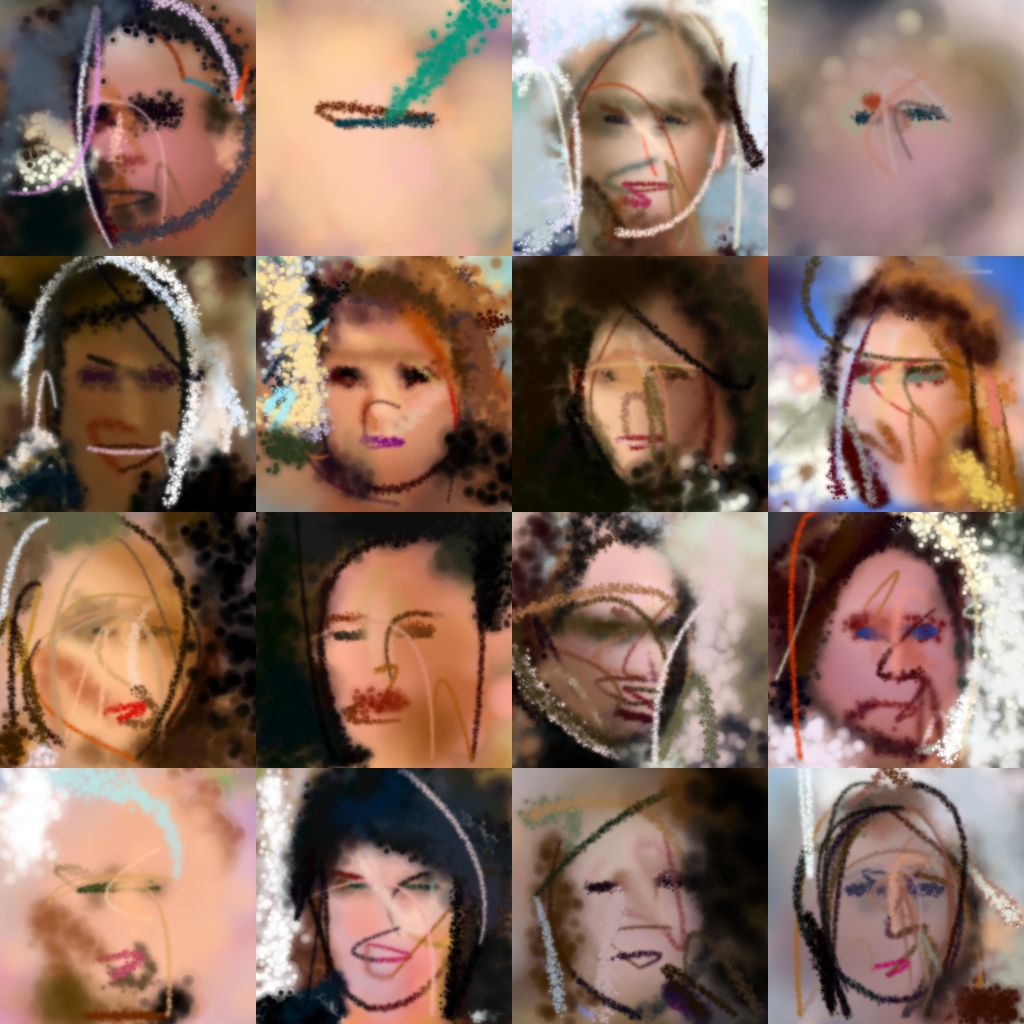}
    \caption{No credit assignment}
\end{subfigure}
\hfill
\begin{subfigure}[b]{0.49\textwidth}
    \includegraphics[width=\textwidth]{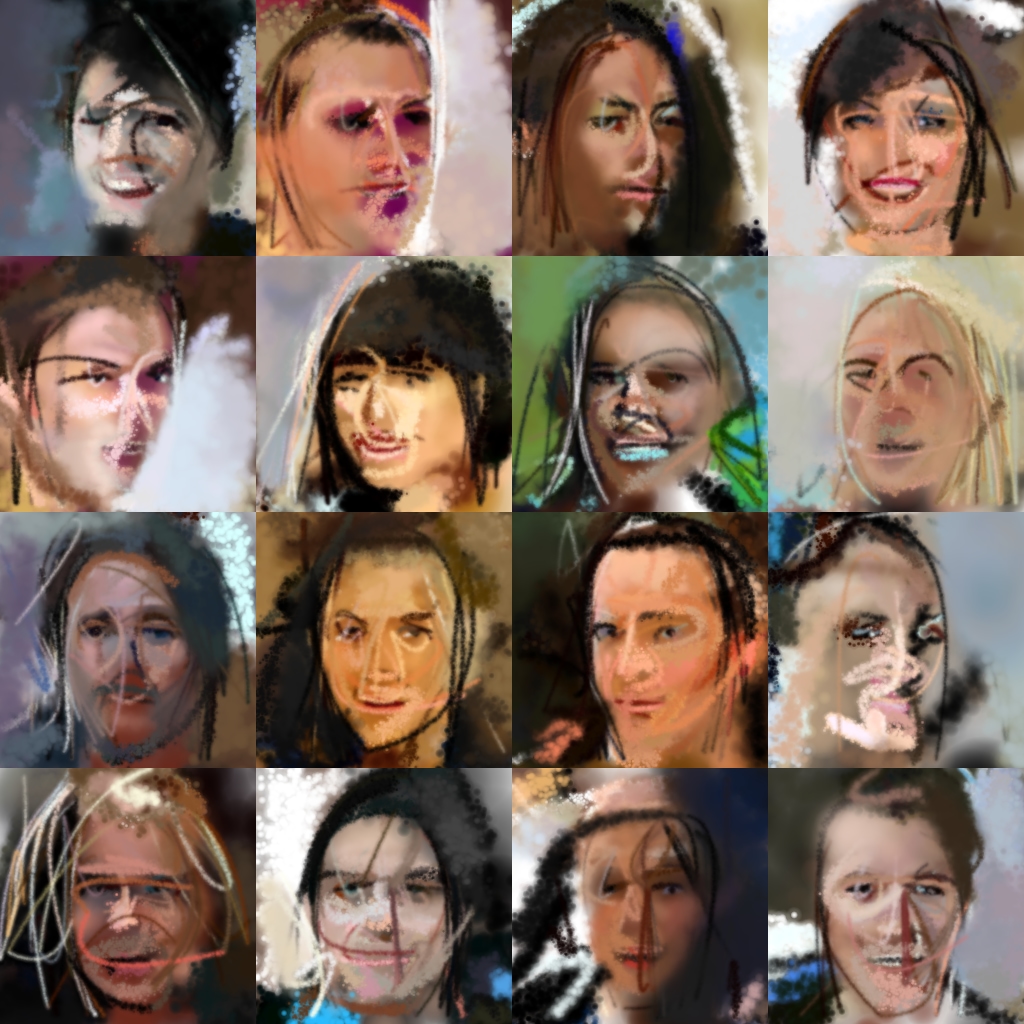}
    \caption{TCA (discount 0.0)}
\end{subfigure}
\caption{\textbf{TCA lets agents make better use of 199 steps.} Random unconditional samples.}
\label{fig:ca_ablation_gen}
\end{figure}

As mentioned in \autoref{sect:temporal_credit_assignment}, agents without TCA often struggle to make good use of long episodes. This can be seen in \autoref{fig:noca_200steps}. It happens even if episode length is gradually increased as a curriculum. The impact on sample quality of adding TCA can be seen in \autoref{fig:ca_ablation_gen}.

Remarkably, we were also able to train good agents with a discount factor of precisely zero, meaning they completely greedily paint one stroke at a time all the way from a blank white canvas to a final image. This suggests that in expectation the discriminator loss monotonically decreases from a blank canvas to a completed image. This works both when training the discriminator to reject intermediate canvases as fakes (in which case there is greater diversity of intermediate canvases the further away you get from a blank canvas) and when only training the discriminator on the final canvas.

\FloatBarrier\inlinesubsection{Complement discriminator ablation}
\label{sect:complement_ablation}

\begin{figure}[H]
\centering
\begin{subfigure}[b]{0.32\textwidth}
    \includegraphics[width=\textwidth]{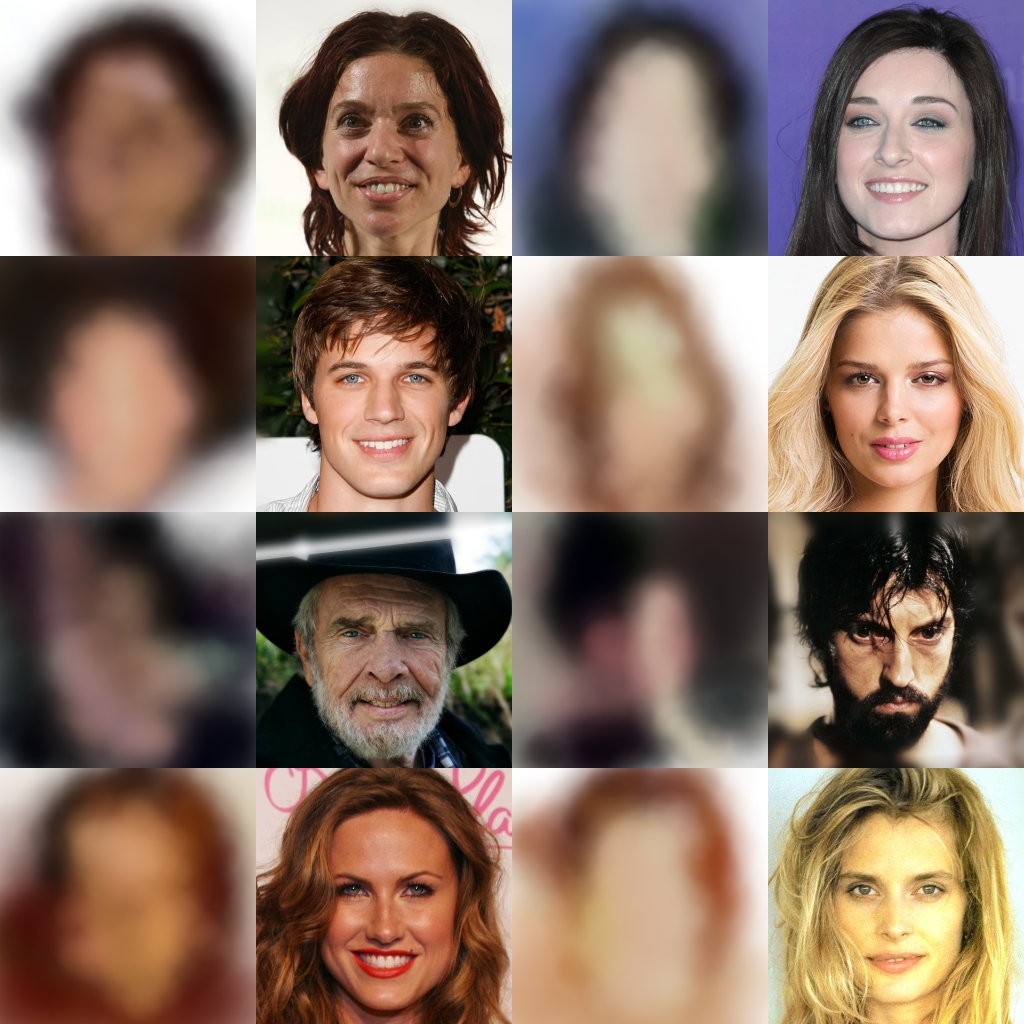}
    \caption{L2 loss}
\end{subfigure}
\hfill
\begin{subfigure}[b]{0.32\textwidth}
    \centering\includegraphics[width=\textwidth/2]{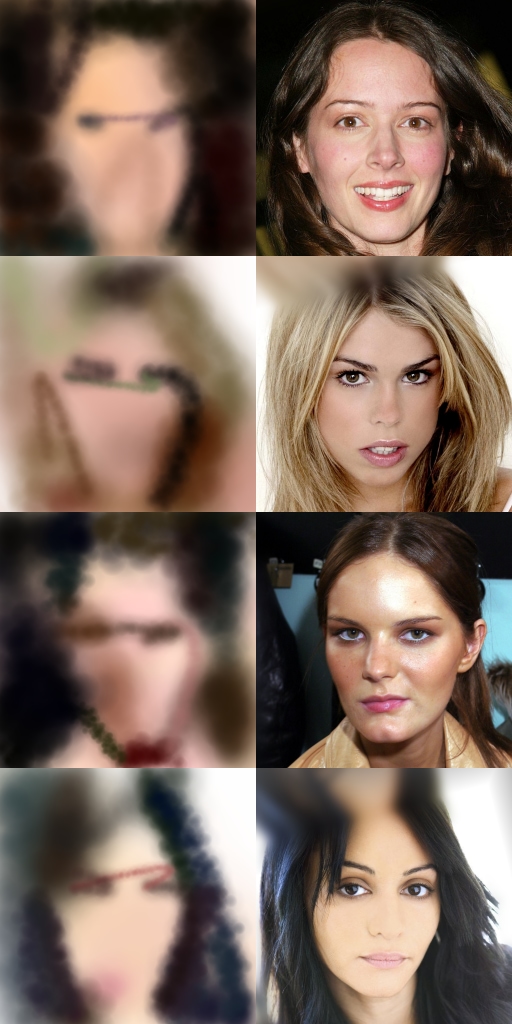}
    \caption{Conditional GAN}
\end{subfigure}
\hfill
\begin{subfigure}[b]{0.32\textwidth}
    \includegraphics[width=\textwidth]{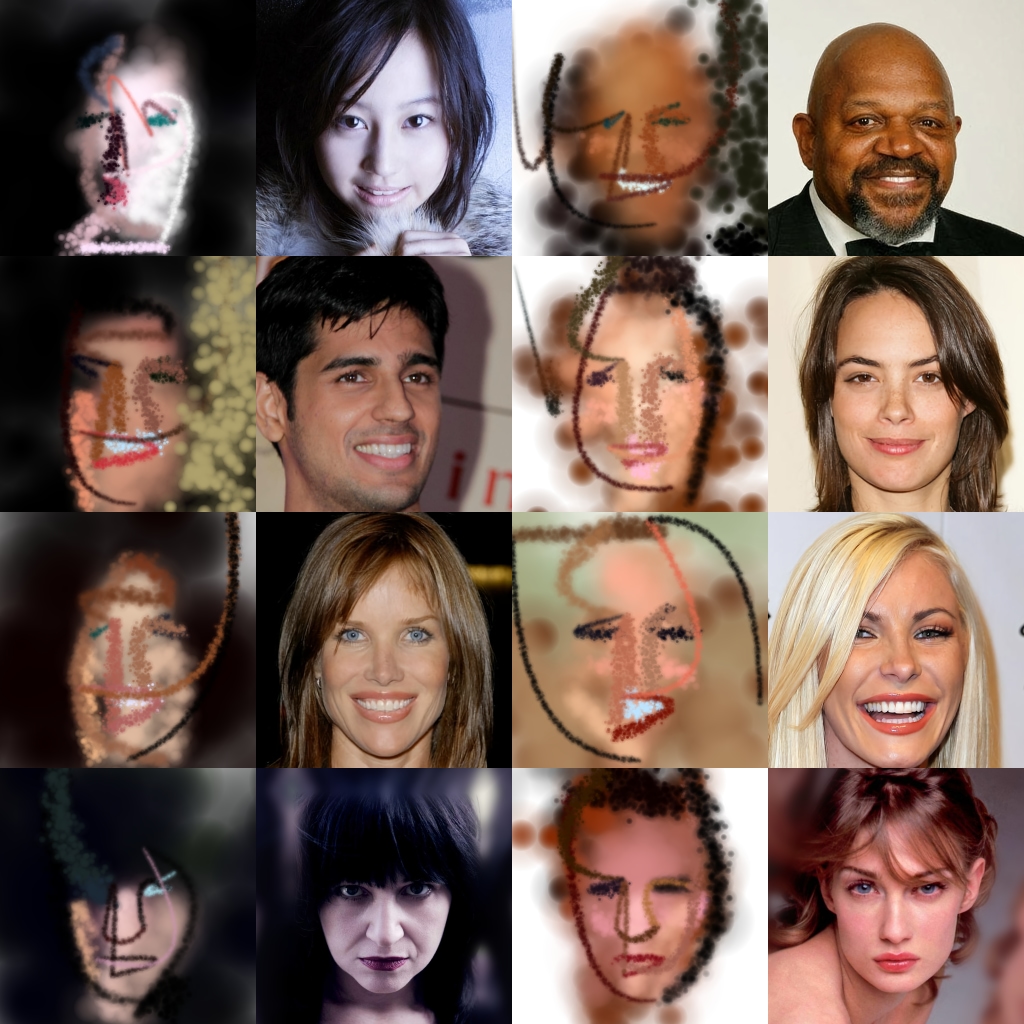}
    \caption{Complement Discriminator}
\end{subfigure}
\caption{\textbf{Types of conditioning 20 step.} The Complement Discriminator gives rise to more interesting reconstructions that are semantically similar rather than similar in pixel space, for example mouths are open wider if the person was smiling widely.}
\label{fig:conditioning_ablation}
\end{figure}

\begin{figure}[H]
\centering
\includegraphics[width=\textwidth]{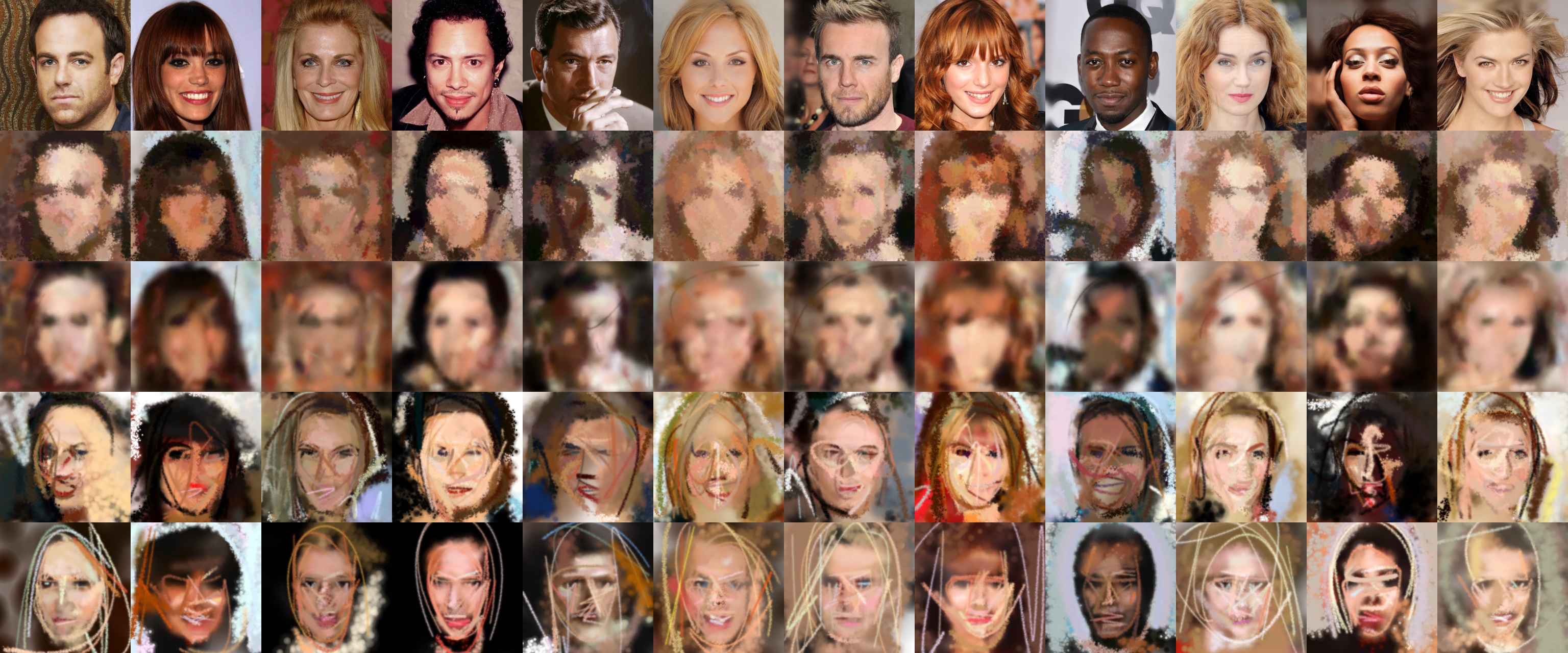}
\caption{\textbf{Types of conditioning 200 step.} Four 200-step agents reconstruct the target images in the top row. The top two agents paint bezier curves and were trained using L2 loss with discount $ \gamma $ of 0 and 0.9 respectively. The bottom two agents were both trained with Complement Discriminator, and paint bezier and spline curves respectively (see \autoref{sect:compound_action_space}).}
\label{fig:conditioning_ablation_ca}
\end{figure}

\autoref{fig:conditioning_ablation} and \autoref{fig:conditioning_ablation_ca} contrast reconstruction with L2 loss or fully-conditioned discriminator against reconstruction with the Complement Discriminator from \autoref{sect:complement_discriminator}.

\inlinesubsection{Action space ablation}
\label{sect:action_space}

\begin{figure}[p]
\centering
\adjincludegraphics[width=\textwidth, trim={0 0 {.16667\width} 0}, clip]{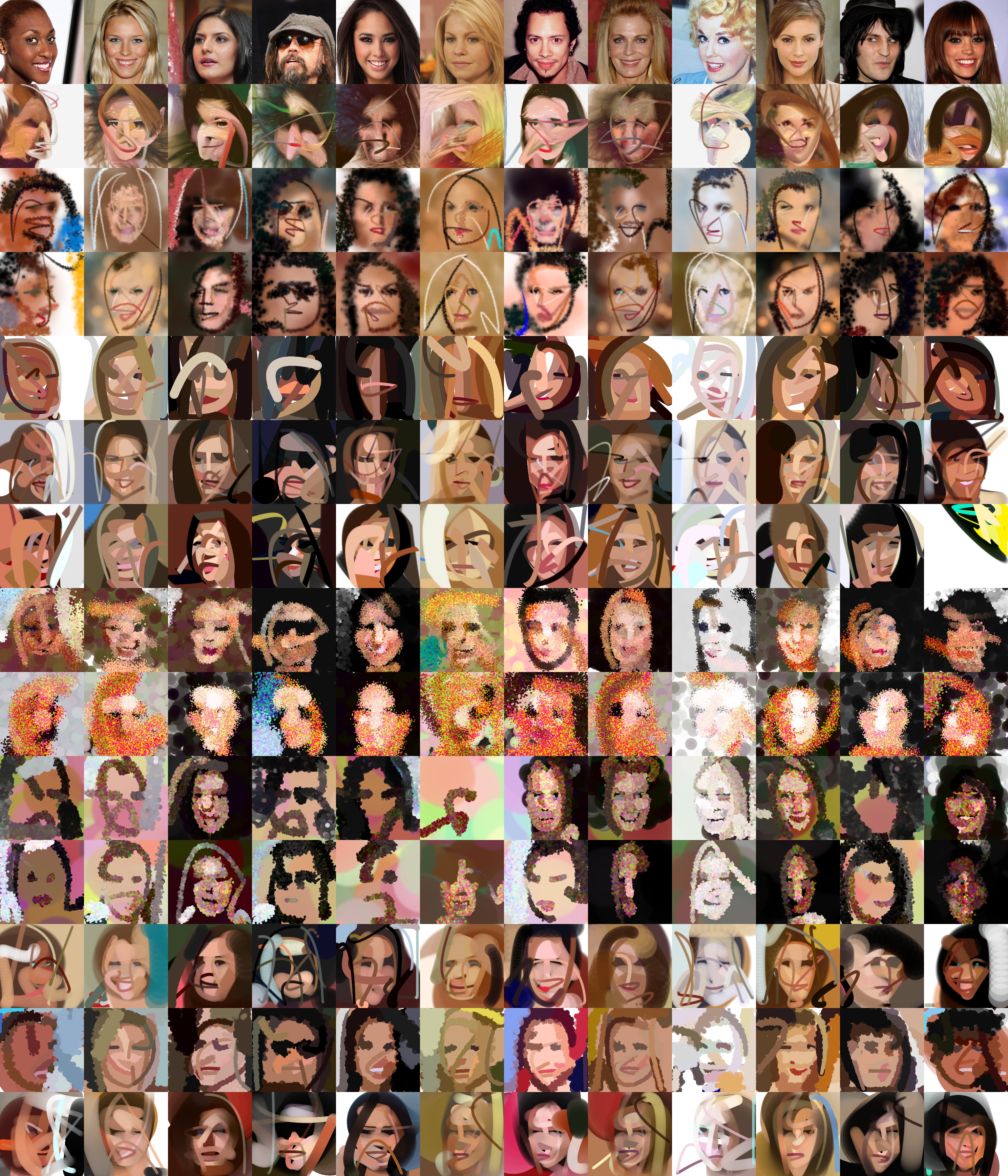}
\caption{\textbf{Training with different brushes leads to very different styles.} Agents with identical architectures perform reconstruction of the same target photos (top row) for 20 steps; only the brush they were trained with varies (except the first row which uses the Fluid Paint renderer, \autoref{sect:comparison_with_model_based}). Continued on next page.}
\label{fig:brush_ablation}
\end{figure}

\begin{figure}[p]
\centering
\adjincludegraphics[width=\textwidth, trim={0 0 {.16667\width} 0}, clip]{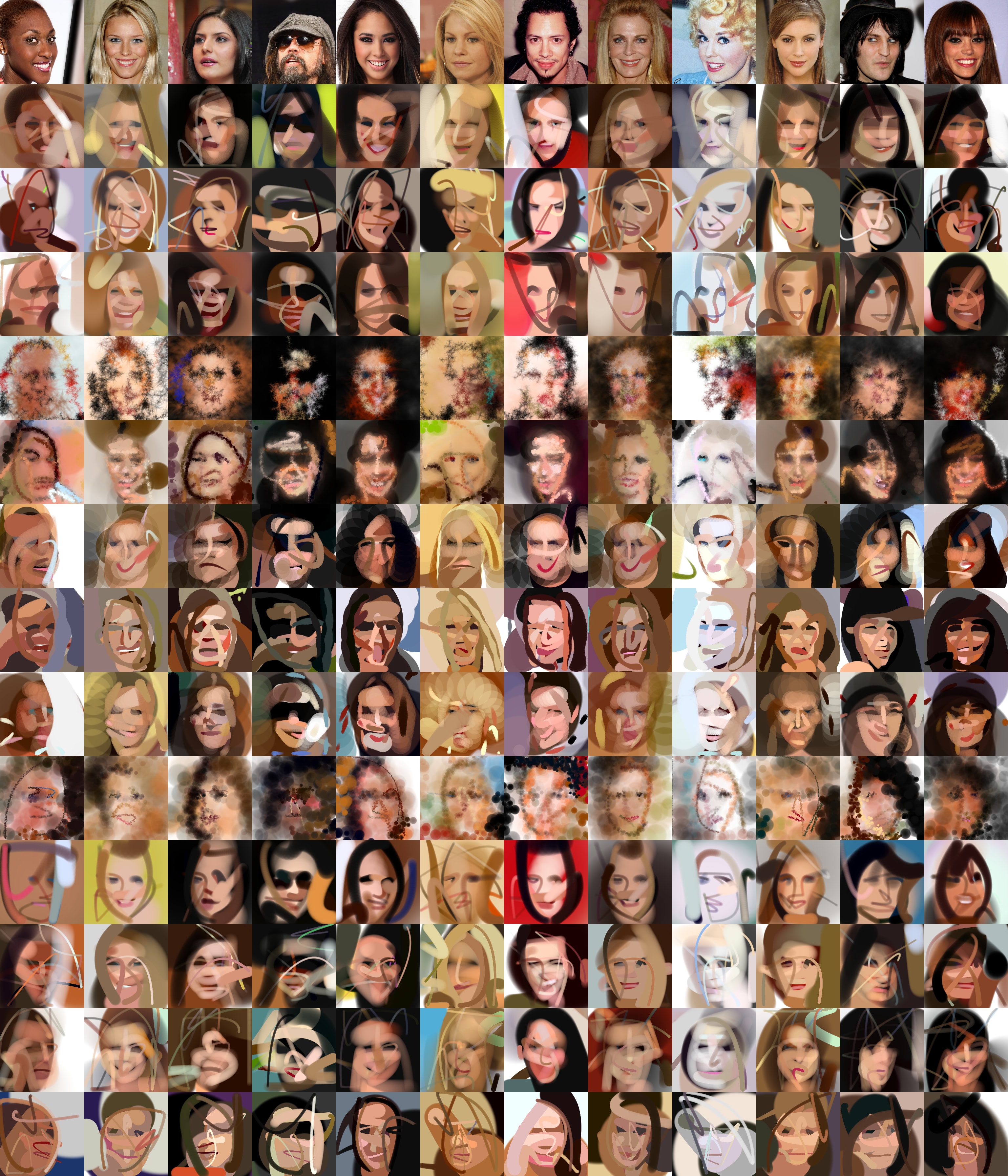}
\caption{\textbf{Training with different brushes leads to very different styles.} Agents with identical architectures perform reconstruction of the same target photos (top row) for 20 steps; only the brush they were trained with varies. Continued from previous page.}
\label{fig:brush_ablation_2}
\end{figure}

When \generativeagentsname learn to operate human software like painting programs, they have a human-interpretable action space, and you can control the model by changing the actions offered by the environment. \autoref{fig:brush_ablation} and \autoref{fig:brush_ablation_2} show how retraining the agent with different brushes leads to very different styles.

\FloatBarrier\section{Comparison with recent model-based approaches to painting}
\label{sect:comparison_with_model_based}

Recently, several papers \citep{frans2018unsupervised, nakano2019neural, zheng2018strokenet, huang2019learningtopaint} proposed to improve sample complexity and stability of \generativeagentsname by replacing the actual rendering simulator with a differentiable neural surrogate. The latter is trained offline to predict how a certain action (usually random) affects the state of the canvas. Although this is a promising avenue for research, we would like to mention several scenarios in which model-free approaches like \oldmodelname and \newmodelname would be more suitable alternatives. 

First, one of the most appealing features of the neural environments is that they allow to train agents by directly backpropagating the gradient coming from the objective of interest (\eg, reconstruction loss or adversarial generator loss). Unfortunately, this means that one has to stick to continuous actions in order to avoid usage of brittle gradient approximators. Continuous actions may be fine for defining locations (\eg, the end point in a B\'ezier curve) but are not appropriate for the situations requiring inherently discrete decisions like whether the agent should lift the brush or add one more control point to a spline. In \autoref{sect:results_omniglot}, we show that \newmodelname performs reasonably well in this latter case.

Secondly, the success of the agent training largely depends on the quality of the neural model of the environment. The simulator used in \cite{ganin2018spiral} and in the experiments discussed so far is arguably easy to learn since new stokes interact with the existing drawing in a fairly straightforward manner. Environments which are highly stochastic or require handling of object occlusions and lighting might pose a challenge for neural network based environment models.

\begin{figure}[h]
\centering
\includegraphics[width=\textwidth]{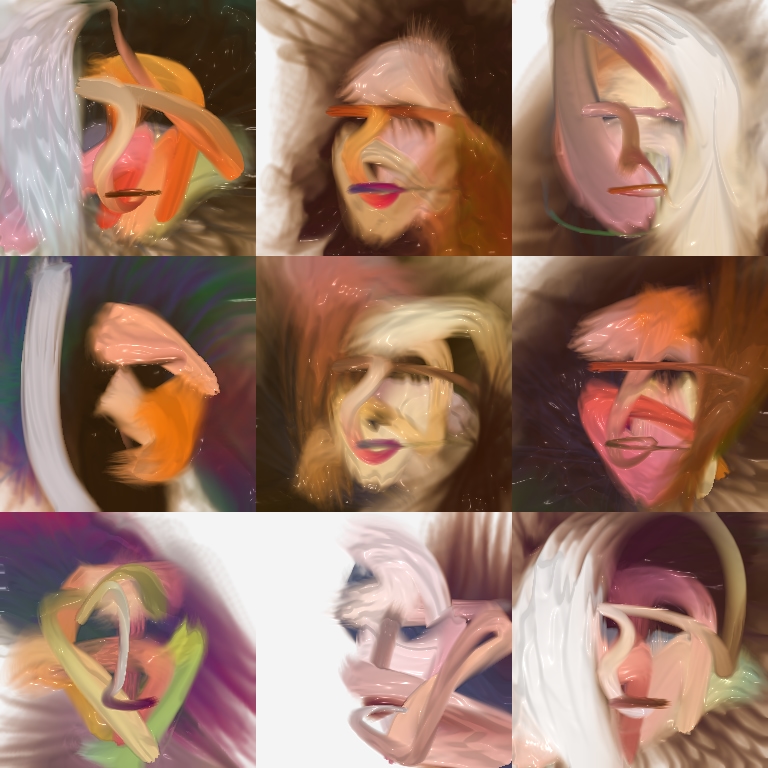}
\caption{\textbf{Selected unconditional 20-step samples.} Notice how these agent can paint hair in just one or two brush strokes, relying on the paint physics to draw many individual strands of hair.}
\label{fig:brush_ablation_fluid}
\end{figure}

Lastly, while in principle possible, designing a model for a simulator with complex dynamics may be a non-trivial task. The majority of the recent works relying on neural renderers assume that the simulator state is fully represented by the appearance of the canvas and therefore only consider non-recurrent state transition models. There is no need for such an assumption in the \oldmodelname framework. We demonstrate this advantage by training our agent in a new painting environment based on Fluid Paint \citep{li2017fluid}. A distinctive feature of this renderer is that under the hood it performs fluid simulation on a grid of cells. The simulation is governed by the Navier-Stokes equations and requires the access to the velocity field of the fluid \citep{fernando2004gpu} which is not directly observable from the drawing. On top of that, unlike MyPaint, Fluid Paint models the behaviour of the brush bristles moving against the canvas surface. Despite having to deal with an arguably more complex setting, \newmodelname managed to not only learn the basic control of tool at hand but also exploit some of its peculiarities in an interesting way (\eg, use of the brush bristles to imitate hair). Please refer to \autoref{fig:brush_ablation_fluid} and the second row of \autoref{fig:brush_ablation} for qualitative results.

\clearpage\section{Additional samples}

\begin{figure}[h]
\centering
\adjincludegraphics[width=\textwidth, trim={0 {.125\width} 0 0}, clip]{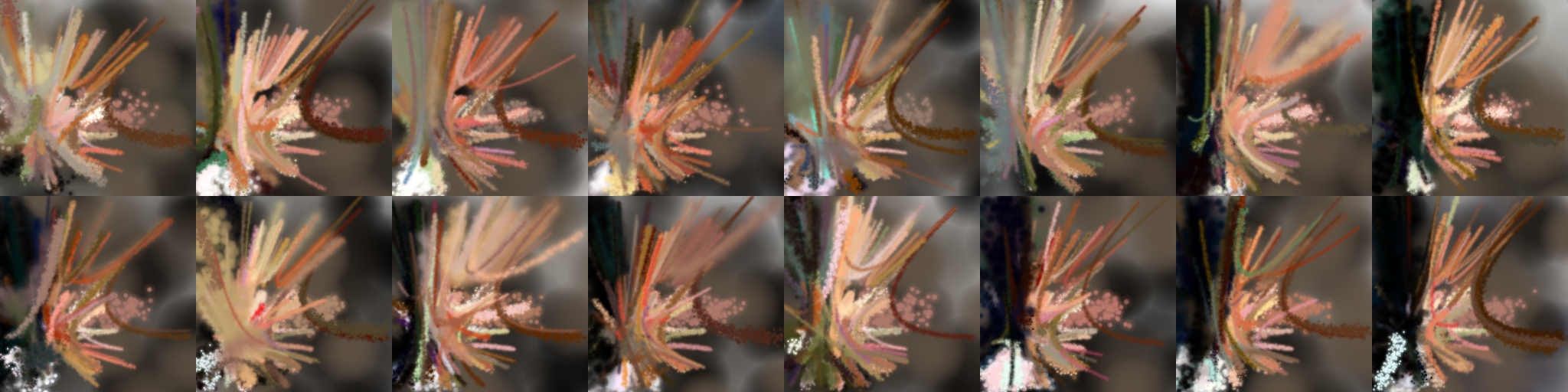}
\caption{\textbf{Radial faces.} This agent learned a rather unusual style, building up faces with arcs originating from a single corner of the image.}
\label{fig:radial_faces}
\end{figure}

\begin{figure}[h]
\centering
\begin{subfigure}[b]{0.5\textwidth}
    \adjincludegraphics[width=\textwidth, trim={0 {.5\width} 0 0}, clip]{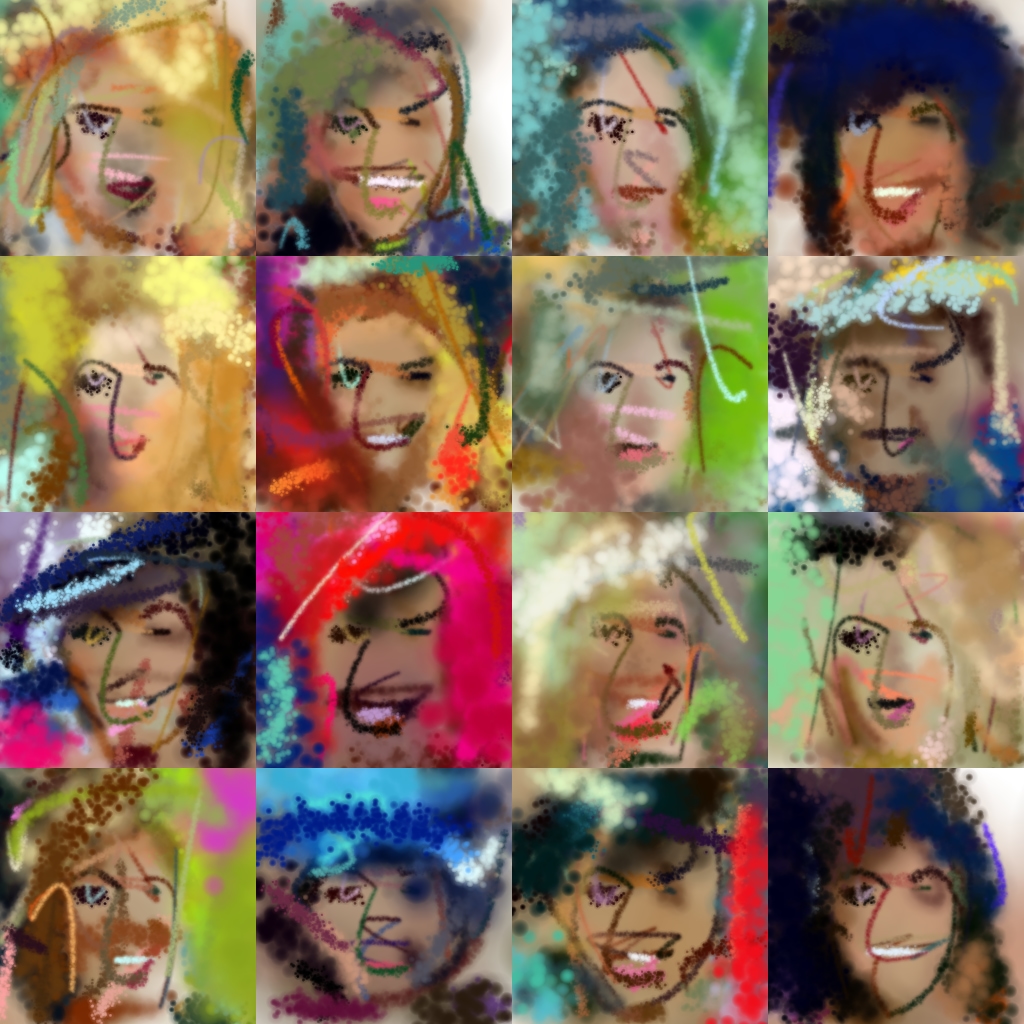}
\end{subfigure}%
\begin{subfigure}[b]{0.5\textwidth}
    \adjincludegraphics[width=\textwidth, trim={0 0 0 {.5\width}}, clip]{im/warhol_grid}
\end{subfigure}
\caption{\textbf{Focus on color.} Agents are occasionally Warholesque in their use of color. Perhaps these are compensating for other agents in the population using drab color palettes?}
\label{fig:warhol_grid}
\end{figure}

\autoref{fig:radial_faces} and \autoref{fig:warhol_grid} show further samples of agents that learned interesting painting styles.

\FloatBarrier\section{Hyperparameters}

The generator learning rates and RL entropy costs were evolved using population based training (\autoref{sect:population_of_generators}), but we found that bad initial values could cause training to fail to converge. For all our experiments each generator sampled initial values of the learning rate from the range (\num{1e-5}, \num{3e-4}) and entropy cost from the range (\num{2e-3}, \num{1e-1}). As in \cite{ganin2018spiral} we trained the discriminator using Adam \citep{kingma2014adam} with a learning rate of \num{1e-4}, $ \beta _1 $ set to $ 0.5 $ and $ \beta _2 $ set to $ 0.999 $. For short episodes we used an RL discount factor $ \gamma $ of 1.0, but for long episodes using temporal credit assignment (\autoref{sect:temporal_credit_assignment}) we found that using values anywhere between 0 (completely greedy) and 0.99 gave good results, on episodes of length up to even 1000 steps. Most of our experiments use unroll lengths of 20 steps; others use unroll lengths of 50 steps with no noticeable performance differences (in particular, longer unroll lengths do not seem to reduce the need for temporal credit assignment (\autoref{sect:temporal_credit_assignment}). We used batch sizes of 64 for both generators and discriminators. Brush strokes were rendered on a 256x256 canvas, but to reduce compute costs the canvas and images from the dataset were downsampled to 64x64 before being fed to the generator or discriminator.

\end{appendices}

\end{document}